%% file: yun.tex
\def\SubmissionVersion{1}
\documentclass{article}
\pdfoutput=1  %
\usepackage{template/fancyhdr,template/natbib,template/iclr2027_conference,times}
\usepackage{amsmath,amssymb,amsthm}
\usepackage{graphicx}
\usepackage{tikz}
\usetikzlibrary{arrows.meta,positioning}
\usepackage{booktabs}
\usepackage{etoolbox}
\usepackage{makecell,multirow,placeins,float}
\usepackage{longtable}
\usepackage{xcolor}
\input{figures/icons/icons.tex}

\usepackage{url}

\title{GEB-Bench: Abstract Structures Told in Many Voices}

\iclrfinalcopy
\author{%
  Tong Zhang\\
  Fudan University\\
  \texttt{tongzhang25@m.fudan.edu.cn}
  \And
  Zhiyuan Shi\\
  \texttt{aron.c.shi@gmail.com}
  \And
  Yun Peng\\
  Fudan University\\
  \texttt{yunpeng4@sigsoft.org}
  \And
  Tao Xie\thanks{Corresponding author.}\\
  Peking University\\
  \texttt{taoxie@pku.edu.cn}
}

\newcommand{\bench}{\textsc{GEB-Bench}}

\ifdefined\ShowReviewComments
\newcommand{\reviewcomment}[1]{\textcolor{red}{[Review: #1]}}
\else
\newcommand{\reviewcomment}[1]{}
\fi

\theoremstyle{definition}
\newtheorem{definition}{Definition}

\ifdefined\SubmissionVersion
\newcommand{\cs}[1]{}
\else
\newcommand{\cs}[1]{\textcolor[RGB]{224,69,86}{[CS: #1]}}
\fi

\ifdefined\SubmissionVersion
\newcommand{\py}[1]{}
\else
\newcommand{\py}[1]{\textcolor{red}{[PY: #1]}}
\fi

\begin{document}
\maketitle
\fancyhead{}
\lhead{}
\rhead{}
\cfoot{\thepage}

\AtBeginEnvironment{tabular}{\fontsize{8}{9.2}\selectfont\setlength{\tabcolsep}{1.7pt}}
\AtBeginEnvironment{tabular*}{\fontsize{8}{9.2}\selectfont\setlength{\tabcolsep}{1.7pt}}
\AtBeginEnvironment{longtable}{\fontsize{8}{9.2}\selectfont\setlength{\tabcolsep}{1.7pt}}

\begin{abstract}
Analogy allows models to apply relational knowledge from one situation to
another. Benchmarks of equivalent representations test different ways of
expressing the same underlying problem, whereas analogy often links
situations involving different objects and content.
To evaluate the analogy ability of models, we introduce \bench{},
a benchmark centered on cross-voice motif matching. A voice presents
a motif as a scene, story, mathematical statement, or programmatic
skeleton. The first task is identification across these voices.
The second is matching scenes to stories or mathematical statements,
and stories to mathematical statements, by their shared motif. A complete construction pipeline generates
items in these voices from the relation specified for each motif, checks the order
and repetition of story units, and assembles paired questions.
We evaluate current models on our benchmark. Results show that matching
performance depends on the target voice. The paired scene-to-story study reveals that
(1) evaluated models still make matching errors when they separately
identify the source and every candidate correctly, and (2) identification
improves more consistently than scene-to-story matching across the
evaluated Qwen model sizes. These findings show that successful motif
identification does not ensure reliable matching across voices.
\end{abstract}

\section{Introduction}
\label{sec:intro}

Humans draw analogies between situations with different objects and
appearances by recognizing similar relationships among their parts. This ability makes knowledge
useful beyond the circumstances in which it was acquired. A relation
learned in one setting can organize an unfamiliar situation and suggest
which aspects deserve attention. Structure-mapping theory describes this
process through systematic relational
correspondences~\citep{gentner1983structure}. Developmental research
identifies comparison as a route to category
abstraction~\citep{gentner1999comparison}, while cognitive approaches to
human-like learning emphasize compositional representations that support
reuse across tasks~\citep{lake2017building}. These accounts motivate a
central question for multimodal models. Can a model carry an identified
motif from one form of representation to another when the surrounding
content changes?

Abstract transfer is the key to analogy. It preserves a relational
pattern across changes in objects, wording, or visual appearance.
We study it through identification and matching. Identification assigns
a motif label to an item. Matching selects an item in another voice
with the same motif. Their relationship is practically
important because an identified motif can provide the basis for a later
comparison. Language models already show strong performance on several
analogy tasks~\citep{webb2023emergent}, and their sensitivity to problem
variants and answer ordering reveals how much the presentation can
matter~\citep{lewis2024robustness}. Evaluating identification alongside
matching offers a way to study how models use familiar relations
when comparing different content.

Different media make relational evidence available in different ways.
An image can communicate a relation through spatial organization,
whereas a text can express it through the order and recurrence of its
parts. The model must select the relevant organization from the details
of each input and relate the resulting descriptions. This creates a
challenge beyond detecting the same object or retrieving a familiar
concept name. Both inputs may provide clear evidence in their own
conventions while sharing little vocabulary or visual appearance.
A useful evaluation should preserve this variation in content and
specify the motif that defines a correct match.

Classical structure-mapping systems illustrate the role of an explicit
relational representation. The Structure-Mapping Engine receives
relational descriptions and constructs structurally consistent
correspondences between them~\citep{falkenhainer1989sme}. Multimodal models
receive images and natural language, so extracting a suitable description
is itself part of the task. This difference suggests an experimentally
useful question. When a model identifies the motif in each item
separately, does it also match those items correctly?
Paired identification and matching questions connect the final answer
to the model's component judgments.

Existing benchmarks provide precise tests for several related questions.
ARC measures induction of transformations from visual grid
problems~\citep{chollet2019arc}, and ConceptARC groups related tasks around
systematically varied concepts~\citep{moskvichev2023conceptarc}.
They both study rule application within a common visual format.
IsoBench compares visual problems with isomorphic textual representations,
including mathematical encodings in relevant
domains~\citep{fu2024isobench}. SEAM pairs semantically equivalent textual
and visual notations in formal domains~\citep{tang2025seam}.
These representation pairs preserve the specified problem or semantic
content. MathVerse varies how information is distributed between text and
diagrams to examine the use of visual mathematical
information~\citep{zhang2025mathverse}. Together, these designs establish
ways to study rule induction, consistency between equivalent encodings
and the information supplied by each modality. Table~\ref{tab:benchmark-comparison}
compares these designs with the shared-motif construction and paired
identification and matching tasks in \bench{}.

Visual analogy benchmarks further clarify the role of the relation being
compared. VASR evaluates analogy completion through changes between
natural-image scenes~\citep{bitton2023vasr}. It asks models to recognize a
transformation that connects a visual pair and find an analogous change
elsewhere. Our question concerns a different connection between
representations. A scene and a story can express a shared pattern through
spatial and narrative organization without depicting corresponding
events. Evaluating this connection requires the evidence in each voice
to be identifiable in its own terms. It also requires measuring whether
the model can complete the match after those individual motifs
have been identified.

\begin{table}[t]
\caption{Benchmark designs for relational reasoning and multimodal comparison.
\bench{} connects different content through shared motifs, generates voices
from relational specifications, and evaluates identification alongside matching.}
\label{tab:benchmark-comparison}
\centering
\input{revision_storyline/tables/benchmark_comparison}
\end{table}

Analogy also connects different situations through a shared organization.
For this setting, preserving an entire problem would remove the content
variation of interest, while relying on broad semantic similarity would
leave the relation to be compared unclear. The benchmark design therefore
needs a relational target that can be realized with different entities
and through different presentation conventions. Scenes and stories
must each contain observable evidence for that target. Otherwise,
agreement could reflect shared subject matter, and disagreement could
reflect an unclear realization. Construction therefore starts from
the relation to be expressed and specifies how it appears in each voice.

We define cross-voice motif matching and introduce \bench{} to evaluate
this task alongside motif identification. We use a motif to represent
the intended relational pattern shared by its assigned items.
Inspired by the cross-media perspective of G\"odel, Escher,
Bach~\citep{hofstadter1979geb}, we use voice to mean a motif representation.
Given a source item, matching selects a candidate in another voice with
the same motif. Scene-to-story and scene-to-mathematical-statement matching cross
modalities; story-to-mathematical-statement matching compares textual voices.
For example, a story expresses its motif through the organization of
its telling. Mathematical statements and skeletons provide compact
references for selected relations. These roles let the benchmark vary
content while keeping the target motif explicit.

\begin{figure}[t]
\centering
\includegraphics[width=\linewidth]{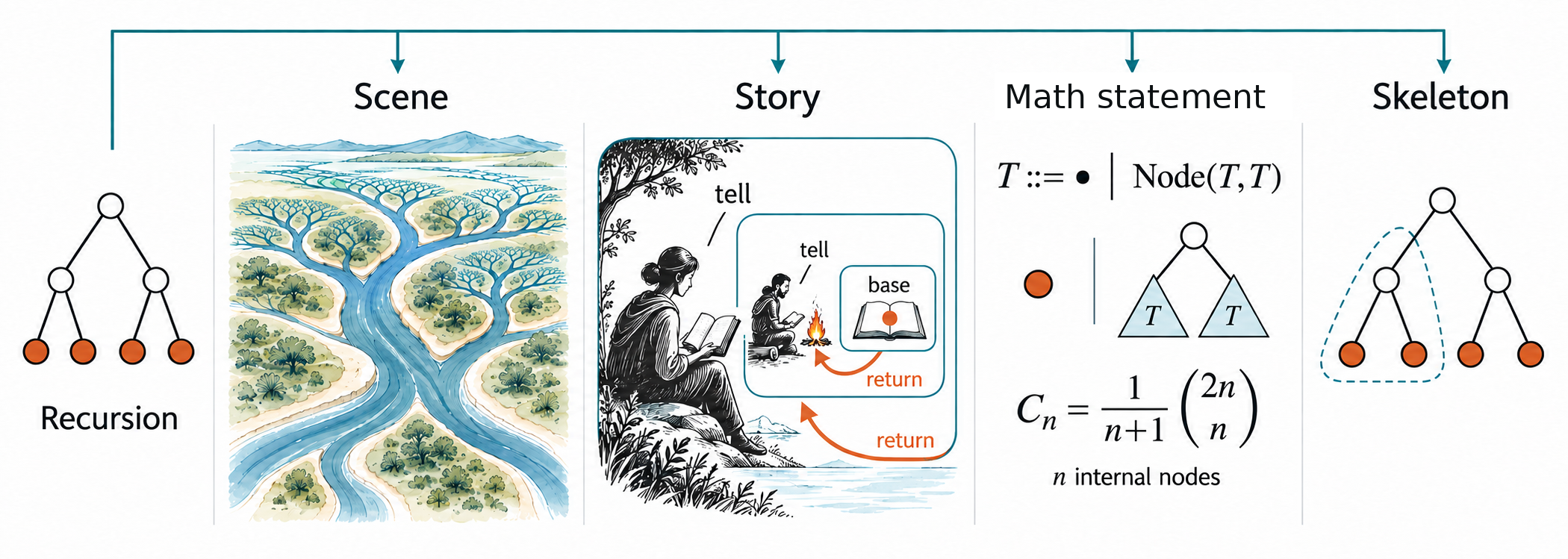}
\caption{A motif across voices. These illustrative redraws present
recursion through a scene, story, mathematical statement, and skeleton. Identification assigns a motif label to an item;
matching selects an item in another voice with the same motif.}
\label{fig:voices}
\end{figure}

Cross-voice motif matching addresses the challenge of comparing
different content by requiring the source and target to share a
specified motif. For the recursion
example in Figure~\ref{fig:voices},
identification asks which relation the scene expresses. Matching asks
which candidate story expresses that relation through the organization
of its telling, despite its different subject matter. The source scene
and candidate motifs are shared across these tasks, linking the final
choice to identification of its components. A complete pipeline generates
items in these voices from motif specifications and checks their form before
assembling the paired questions.

We evaluate current models on motif identification and cross-voice matching
tasks. Matching performance varies across target voices. Results on
scene-to-story matching reveal that models make errors even on questions
where they correctly identify the source and every candidate separately.
Furthermore, results across Qwen model sizes show that improvements in
identification do not consistently translate into improvements in
scene-to-story matching. These findings locate a persistent difficulty
in matching familiar motifs across different content.

\paragraph{Contributions.} We summarize our key contributions as follows.
\begin{itemize}
\item We define cross-voice motif matching and introduce \bench{} to
evaluate whether models connect different content through a shared motif,
alongside identification of the items being compared.
\item We build a complete construction pipeline connecting motif
specifications, item generation, structural checks, and model review
to paired evaluation questions.
\item We show that models can correctly identify the source and every
candidate yet fail to match them, and that identification improves more
consistently than scene-to-story matching across the evaluated Qwen models.
\end{itemize}

\section{Related Work}\label{sec:related}

\paragraph{Analogy and abstract reasoning.}
Structure-mapping theory explains analogy through systematic relations
among parts~\citep{gentner1983structure}; the Structure-Mapping Engine
constructs correspondences from explicit relational
descriptions~\citep{falkenhainer1989sme}. Human comparison supports
abstraction~\citep{gentner1999comparison}, and language-model analogy
performance varies with problem variants and answer
order~\citep{webb2023emergent,lewis2024robustness}.
Visual analogy benchmarks study changes within a visual
format~\citep{hill2019learning,teney2020vprom,bitton2023vasr}.
ARC and ConceptARC test rule induction and concept
reuse~\citep{chollet2019arc,moskvichev2023conceptarc}.
Raven-style and Bongard tasks extend concept evaluation across synthetic
and natural images~\citep{zhang2019raven,barrett2018measuring,bongard1970,nie2020bongardlogo,jiang2022bongardhoi,wu2024bongardopenworld}.
SCAN and COGS evaluate new combinations through dedicated
splits~\citep{lake2018scan,kim2020cogs}. \bench{} evaluates how models identify
and match a shared motif across different content and voices.

\paragraph{Multimodal evaluation.}
MMMU, MMBench and BLINK assess broad knowledge and
perception~\citep{yue2024mmmu,liu2024mmbench,fu2024blink}; MathVista and
elementary visual-relation tests probe more specific
skills~\citep{lu2024mathvista,rahmanzadehgervi2024blind}.
IsoBench and SEAM compare equivalent problems or content across
representations~\citep{fu2024isobench,tang2025seam}.
Winoground tests compositional image--text
matching~\citep{thrush2022winoground}.
MathVerse varies the distribution of information between text and
diagrams~\citep{zhang2025mathverse}, while SugarCrepe and MMMU-Pro examine
candidate cues and language-only
shortcuts~\citep{hsieh2023sugarcrepe,yue2025mmmupro}.
These studies motivate pairing task outcomes with input-specific
measurements. Our paired questions connect scene-to-story matching
to independent motif identification of the source and candidates.

\paragraph{Benchmark construction.}
Instruction-following evaluation and COLLIE use programmatically checkable
generation constraints~\citep{zhou2023instruction,yao2023collie}.
\bench{} applies such checks to the organization of a story, linking
its specified motif to the item presented during evaluation.
Research on model judges documents the behavior and biases of these
assessments~\citep{zheng2023judging}. We combine executable form checks
with model review and human evaluation of the relations expressed by
the generated items.

\input{revision_storyline/sections/method}
\input{revision_storyline/sections/results}

\section{Conclusion}
\label{sec:conclusion}
\bench{} evaluates motif identification and matching across voices.
Correct identification of the source and every candidate still leaves
substantial matching errors. Across the evaluated Qwen model sizes,
identification improves more consistently than scene-to-story matching;
matching the same scenes to mathematical statements is also generally
easier than matching them to stories. Together, these findings identify
matching across voices as a challenge beyond motif identification.
Future work can extend the library and test whether corresponding
transformations remain aligned as scenes and stories change.

\clearpage
\section*{AI Use Statement}
In this work, we used generative AI tools for language polishing and coding
assistance. AI-assisted generation of benchmark items is described in
Appendix~\ref{app:repro}. The authors reviewed the AI-assisted text and code.
We take responsibility for the final content of this work, including
the code, benchmark items, analyses, and claims produced with the aid of
generative AI.

\section*{Ethics Statement}
The benchmark uses generated scenes and stories and programmatically
rendered skeletons. Human annotations assess these benchmark items, and
we report aggregate agreement under the annotation protocol described
in Section~\ref{app:human-evaluation}.

\section*{Reproducibility Statement}
Appendices~\ref{app:repro}, \ref{app:paired-experiments} and \ref{app:prompts}
document the construction specifications, evaluation prompts, model
settings and scoring rules. The research package contains the benchmark
inputs, evaluation code, model responses and analysis scripts used to
compute the reported results.

\begingroup\small
\bibliography{revision_storyline/references}
\bibliographystyle{template/iclr2027_conference}
\endgroup

\clearpage
\appendix
\raggedbottom
\renewcommand{\topfraction}{0.95}
\renewcommand{\bottomfraction}{0.90}
\renewcommand{\textfraction}{0.05}
\renewcommand{\floatpagefraction}{0.65}
\setcounter{topnumber}{5}
\setcounter{bottomnumber}{4}
\setcounter{totalnumber}{8}
\makeatletter
\setlength{\@fptop}{0pt}
\setlength{\@fpsep}{10pt plus 1fil}
\setlength{\@fpbot}{0pt plus 1fil}
\makeatother
\FloatBarrier
\section{Motif gallery}

\begin{figure}[H]
\centering
\includegraphics[width=0.98\linewidth]{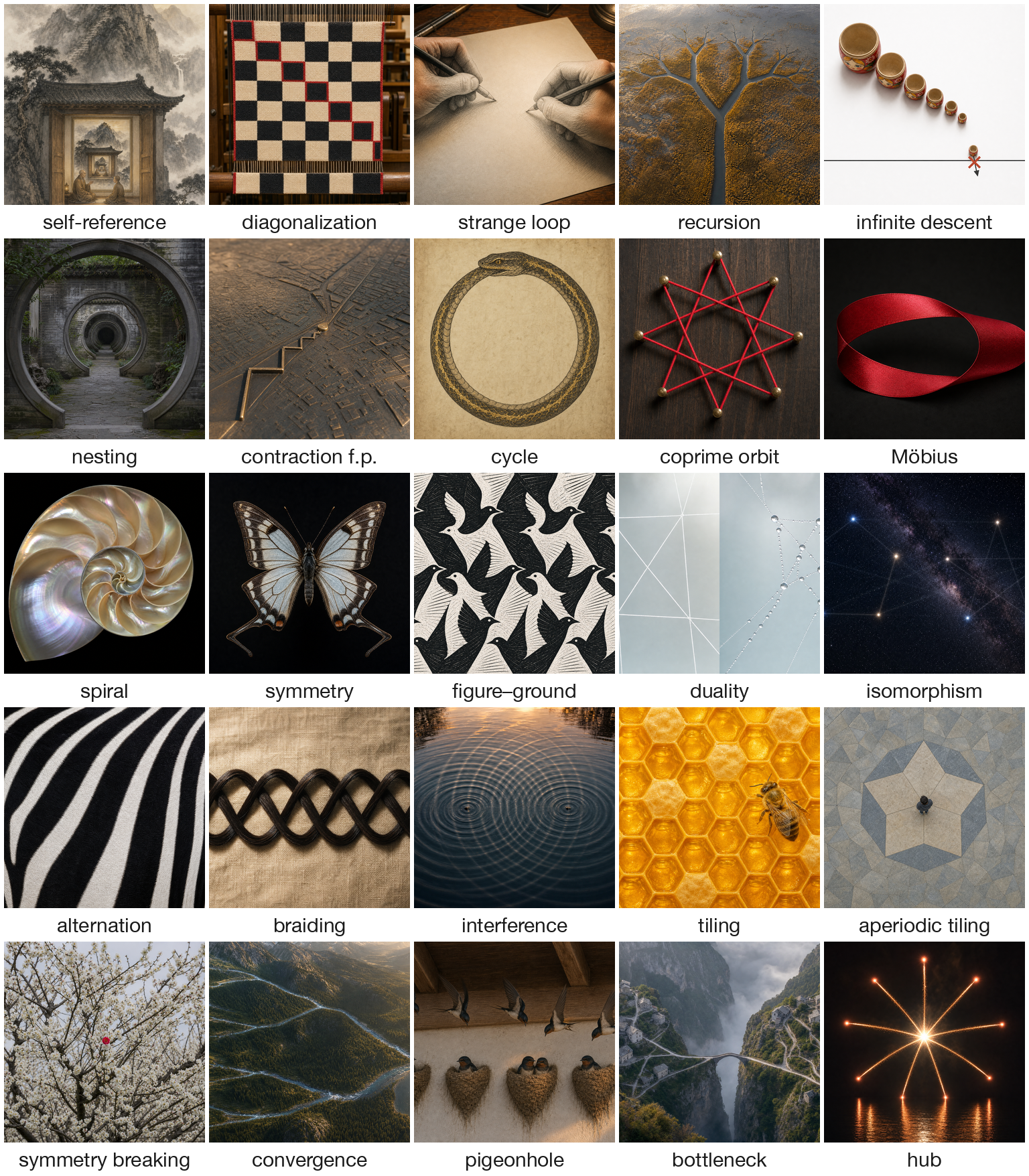}
\caption{The $25$ motifs of \bench{}, each shown through one generated scene.
In reading order, self-reference, diagonalization, strange loop, recursion, infinite
descent; nesting, contraction, cycle, coprime orbit, M\"obius; spiral,
symmetry, figure and ground, duality, isomorphism; alternation, braiding,
interference, tiling, aperiodic tiling; symmetry breaking, convergence,
pigeonhole, bottleneck, hub. Each image is a scene item with its
assigned motif, with varied visual settings, media and palettes.}
\label{fig:gallery}
\end{figure}

\input{revision_storyline/sections/appendix}

\end{document}

%% file: figures/icons/icons.tex
\providecommand{\modelicon}[2]{\raisebox{-0.12ex}{\includegraphics[width=0.95em,height=0.9em,keepaspectratio]{figures/icons/#1.pdf}}\hspace{0.3em}#2}
\providecommand{\vendorheading}[1]{{\fontsize{8}{9}\selectfont\color{black!65}#1}}
\providecommand{\tablebestlegend}{Bold marks the highest accuracy among displayed models in the same evaluation scope, including ties; comparisons use unrounded scores.}

%% file: revision_storyline/tables/benchmark_comparison.tex
\begingroup
\renewcommand{\arraystretch}{1.12}
\begin{tabular}{@{}>{\raggedright\arraybackslash}p{0.20\linewidth}>{\raggedright\arraybackslash}p{0.235\linewidth}>{\raggedright\arraybackslash}p{0.235\linewidth}>{\raggedright\arraybackslash}p{0.29\linewidth}@{}}
\toprule
Benchmark & Relation compared & Input construction & Evaluation focus \\
\midrule
ARC / ConceptARC\newline\citep{chollet2019arc,moskvichev2023conceptarc} & Grid transformation & Grid pairs and concept groups & Rule induction and application \\
\addlinespace[3pt]
VASR\newline\citep{bitton2023vasr} & Analogous changes between scenes & Situation annotations and image filtering & Visual analogy completion \\
\addlinespace[3pt]
IsoBench\newline\citep{fu2024isobench} & Same problem across encodings & Isomorphic visual and textual inputs & Task accuracy across representations \\
\addlinespace[3pt]
SEAM\newline\citep{tang2025seam} & Same content across formal notations & Paired textual and visual notations & Accuracy and cross-modal answer agreement \\
\addlinespace[3pt]
MathVerse\newline\citep{zhang2025mathverse} & Same mathematical problem & Text and diagram information varied & Answer accuracy and reasoning-step evaluation \\
\midrule
\textbf{\bench{}} & Motif across content and voices & Motif-to-voice generation with form checks & Paired identification and cross-voice matching \\
\bottomrule
\end{tabular}
\endgroup

%% file: revision_storyline/sections/method.tex
\section{Method}\label{sec:library}\label{sec:tasks}

\begin{figure}[t]
\centering
\includegraphics[width=\linewidth]{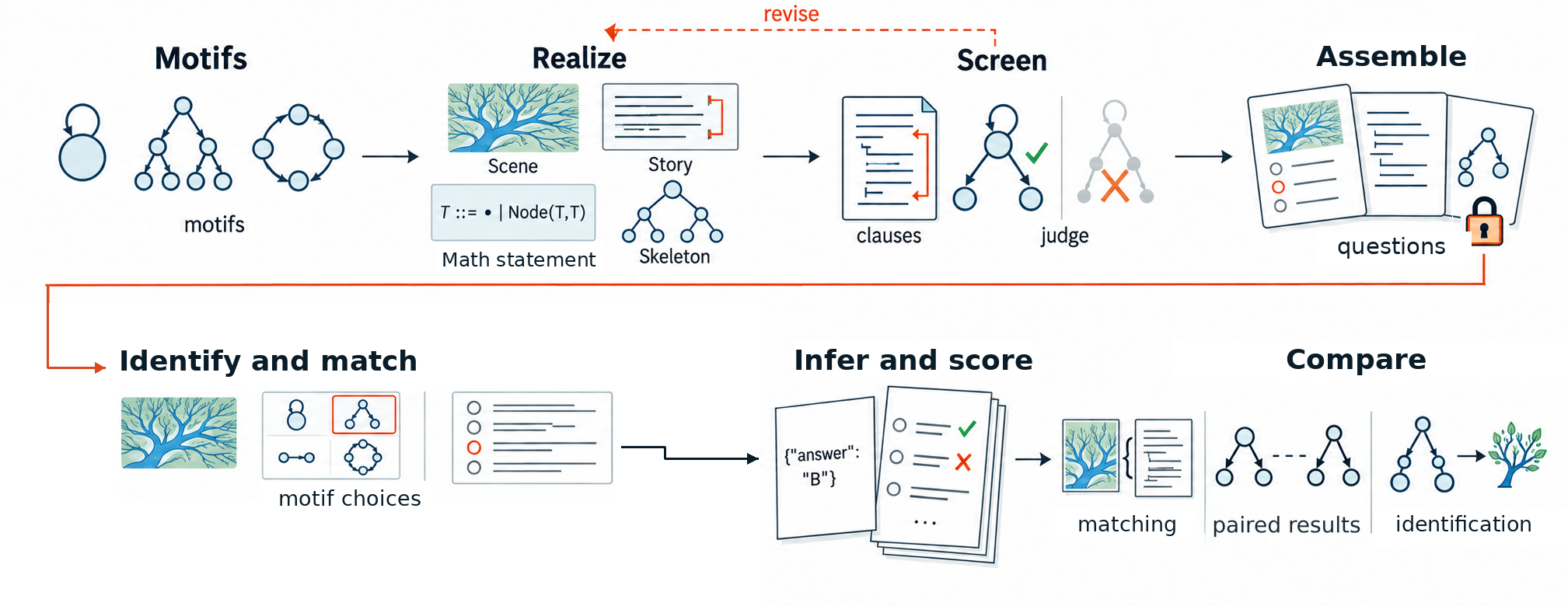}
\caption{Construction and evaluation workflow. Motif specifications guide
item generation in each voice. Structural checks and model-assisted screening
guide revisions before questions are assembled. Identification assigns a motif label to each item; matching selects
an item in another voice with the same motif. Paired results connect
the model's independent identification answers to its matching answer.}
\label{fig:workflow}
\end{figure}

Figure~\ref{fig:workflow} follows a common relational target from
construction to evaluation. The upper path generates items in each voice.
The lower path connects identification to matching. Paired comparisons
reuse the same source items and candidate identities to examine
whether the model can match items whose motifs it identifies independently.

\subsection{Motifs}

Matching across different content requires a motif that remains
identifiable when the objects and subject matter change. We specify this
relation before generating either member of a pair, giving their intended
match a common motif.

\begin{definition}[Motifs and voices]\label{def:motif-cat}
Let $\mathcal M$ be the motif library and
$v\in\{S,T,H,K\}$ denote the scene, story, mathematical statement, and skeleton voices.
A motif $m\in\mathcal M$ specifies a relation among parts; a voice is a
form of presenting that motif. The construction $x=F_v(m;\xi)$ produces
an item $x$ with intended motif $m$ and presentation parameters $\xi$.
\end{definition}

The catalog of named relational concepts and mathematical associations in
Table~\ref{tab:catalog} supplies the motif labels. Each construction
specification describes the motif and how an item expresses it in
each voice. For stories, it prescribes a form device governing how the
telling is organized. A recursive device, for example, specifies nested
units, terminating units, and returns to enclosing units. The relation
thus appears in the story's organization while its subject matter can
vary. Presentation changes preserve the specified relation. For a pigeonhole
construction, for example, varying the number of objects is allowed only
while there remain more objects than available locations.

\subsection{Realize}

To test matching across voices, we give the voices
different construction roles. Scenes and stories provide varied visual
and narrative content, while skeletons and mathematical statements serve
as compact references. Coordinate-level recipes produce skeletons.
Scenes are generated from textual descriptions or by editing skeleton
images, and stories are generated under their specified form devices.
Mathematical statements provide definitions or attributed associations
with the motif.

The English-primary collection provides multiple scene and story
items to vary content. Each motif has a mathematical statement and
some skeleton renderings, reflecting their role as compact
references. Identification and matching use this shared motif library.
The voice inventories and their class distributions are listed in
Table~\ref{tab:motif-counts}.

\subsection{Screen}

The specified relation must be expressed in the generated item.
We check selected form properties with executable predicates, following
work on constrained generation and instruction
following~\citep{yao2023collie,zhou2023instruction}.
For a mirror-symmetry device, the predicate compares paired word-token
lists around the central line under specified normalization rules.
Failed construction clauses guide repair or regeneration, as the return
arrow in Figure~\ref{fig:workflow} indicates.

Model review checks the depicted or narrated relation against
its specification. Together, the predicates and model review guide
construction at the level of textual form and relational content.
Section~\ref{app:label-assessment} reports a separate model review of the
assembled collection.

\subsection{Assemble}\label{sec:assembly}

Reproducible comparisons require fixed question components. After
construction and model review, we remove duplicate story bodies and assemble
questions that preserve source identities, complete candidate texts, and
candidate order. Candidate stories include their titles and full bodies.
A question combines a source item, candidates, and instructions; a task
defines what the question asks the model to predict. Each question carries
its intended answer and input specification.

Identification and matching pairs share ordered candidate motif
identities, allowing their outcomes to be compared on the same source.

\subsection{Identify and match}

A matching question combines source interpretation with candidate
selection. The identification and matching module measures these demands separately
and relates them through shared inputs.

\begin{definition}[Identification and matching]\label{def:object-layer}
For item $x$ in voice $v$ with intended motif $m$ and candidate motif
menu $C\subseteq\mathcal M$ containing $m$, identification returns
$R_v(x;C)\in C\cup\{\bot\}$, where $\bot$ denotes an invalid or absent
predicted label. Its correctness is
\begin{equation}
a(x;C)=\mathbf 1[R_v(x;C)=m].
\label{eq:rightinv}
\end{equation}
A matching question presents a source item $x$ in voice $u$ and an ordered list
$Y=(y_1,\ldots,y_K)$ of $K$ candidate items in voice $v$.
The predicted index $D(x,Y)$ is scored against the unique candidate
index $j^*$ whose intended motif is $m$.
\end{definition}

Paired scene identification uses the same ordered motif identities
as scene-to-story matching. It supplies each motif's name and description
in place of the complete story expressing it.
Matching pairs scenes with stories or mathematical statements, and
stories with mathematical statements. Skeletons are used for motif
identification. The complete task inventory appears in the appendix.

\subsection{Infer and score}

Each response is scored against the intended motif or candidate choice.
Accuracy is the proportion of completed questions answered correctly.
Refusals and responses without a recoverable answer count as incorrect.
Task-level scores separate motif identification from matching across
voices; average accuracy gives equal weight to each task. Scoring rules are specified in Section~\ref{app:answer-scoring}.

\subsection{Compare}

The final stage compares identification with matching on paired
questions. Independent identification answers define the subset where
the source and every candidate receive their intended motif labels.
Matching accuracy on this subset measures how often those successful
component judgments coincide with a correct match. We also compare
task-level performance across model sizes and target voices.

%% file: revision_storyline/sections/results.tex
\section{Experiments}\label{sec:settings}\label{sec:results}

We organize the evaluation around a central question. How well do models use
identified motifs to match items across voices? We first compare
performance across the full model roster, then examine identification and
matching on shared scene-to-story questions. The paired study evaluates
GPT-5.5, GPT-5.2, Gemini 3.8 Flash and Gemini 2.5 Flash, covering paired versions from each model family.

\subsection{Evaluation}\label{sec:voice-gradient}\label{sec:decomposition}

For model $q$, let $a_{q,t}=c_{q,t}/n_{q,t}$ be the accuracy on task $t$,
where $c_{q,t}$ is the number of correct answers and $n_{q,t}$ the number
of completed questions. Average accuracy gives equal weight to the
applicable tasks $\mathcal T_q$,
\begin{equation}
\bar a_q=\frac{1}{|\mathcal T_q|}\sum_{t\in\mathcal T_q}a_{q,t}.
\label{eq:macro}
\end{equation}
All rates and differences are calculated before rounding. Comparisons
use the same completed questions for each model.

The paired study follows the question assembly in Section~\ref{sec:assembly}.
It compares scene identification, individual story identification, and
direct scene-to-story matching. We compare the same model's identification
answers and matching answers on these linked questions.

\subsection{Model comparison}

\begin{figure}[!t]
\centering
\includegraphics[width=\linewidth]{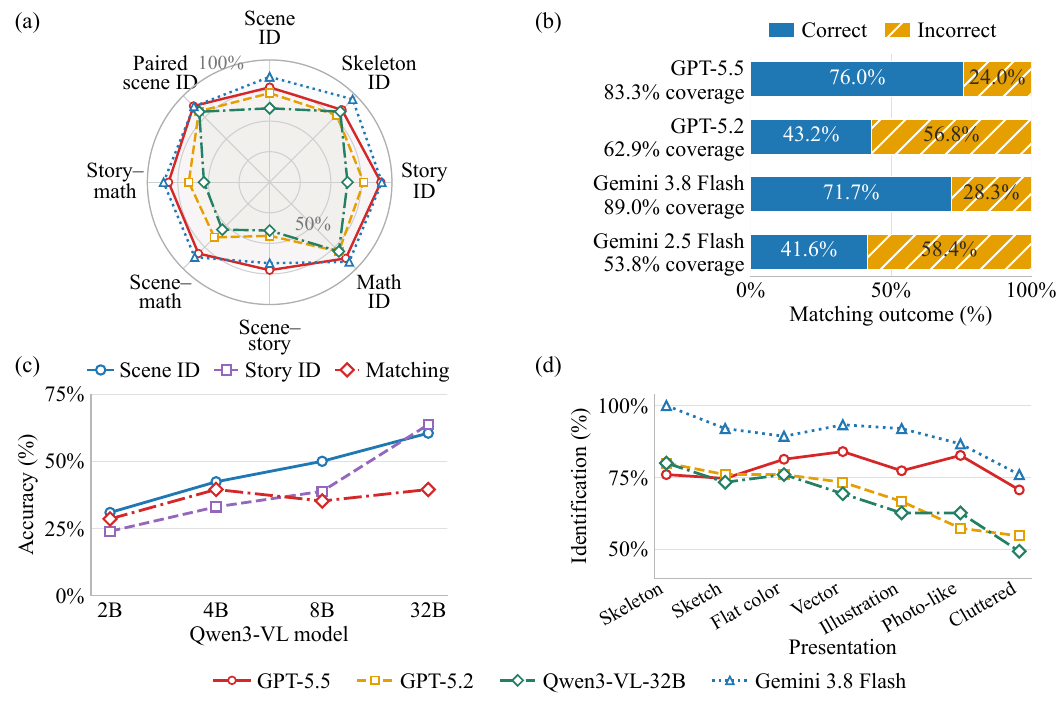}
\caption{Identification and matching across models and presentations.
(a) Primary-task radar profiles.
(b) Matching on each model's subset with correct source and
all-candidate identification; coverage denotes the retained percentage.
(c) Qwen3-VL model-size trends.
(d) Motif-averaged identification across visual presentations.
Model colors are shared by (a,d). All rates are percentages.}
\label{fig:main-findings}
\end{figure}

Table~\ref{tab:main} and Figure~\ref{fig:main-findings}a show model task
profiles. Gemini 3.8 Flash leads in average accuracy, while GPT-5.5 leads
scene-to-story matching. A possible explanation is that motif
identification and comparison across candidate stories impose different
demands.

\begin{table}[t]
\caption{Accuracy (\%) across the primary tasks. Identification uses the full
motif menu. Paired scene identification shares the source scene and
ordered candidate motifs with scene-to-story matching.
Sc denotes scene, St denotes story, Math denotes mathematical
statement, and ID denotes identification. Avg.\ equally weights the primary tasks. Models are grouped by
provider and ordered by this average. \tablebestlegend{}}
\label{tab:main}
\centering\small
\input{revision_storyline/tables/main_results.tex}
\end{table}

\begin{figure}[t]
\centering
\includegraphics[width=\linewidth]{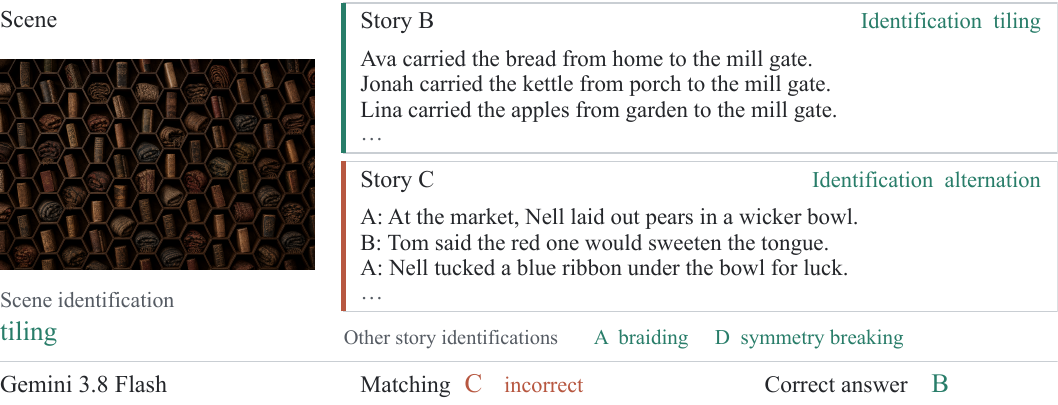}
\caption{An evaluated scene-to-story question. The source scene and
verbatim excerpts from stories B and C show repeated spatial units,
a repeated sentence template, and alternating speakers.
Gemini 3.8 Flash identifies the scene and all candidate motifs correctly,
yet selects C instead of the correct answer B. The other candidate
stories are omitted; their predicted labels are shown.}
\label{fig:identification-example}
\end{figure}

\subsection{Identification and matching}
\label{sec:mapping}\label{sec:adversarial}

\textbf{Models still fail to match inputs whose motifs they identify correctly.}
Matching requires distinguishing the target story from every
alternative. We identify the source scene and every candidate story
separately, and examine direct matching on the questions where
all these predictions are correct. These questions connect the model's component judgments to its matching
answer.

Let $G_i$ equal one when the source and all candidate stories in question
$i$ receive their intended labels, and zero otherwise. If $b_i$ indicates
a correct direct match, the matching accuracy on this subset and its
coverage are
\begin{equation}
A_{\mathrm{identified}}=
\frac{\sum_iG_i b_i}{\sum_iG_i},\qquad
C_{\mathrm{identified}}=\frac{\sum_iG_i}{N}.
\label{eq:conditional}
\end{equation}
Coverage records how much of the full paired pool the comparison uses.
All models complete the same paired pool. Each model defines its own
subset from its independent identification answers.

As shown in Figure~\ref{fig:main-findings}b, correct identification
of the source and every candidate still leaves substantial matching errors for both
newer and earlier models. Coverage is reported for every model because the questions satisfying
this condition differ across models. The result concerns the contrast
within each model between its individual labels and its direct choices.

This pattern is informative because the candidate stories express
different relations. On the selected questions, the model has already
identified those differences and the relation shared by the source
and target. A correct match can therefore be obtained from the component judgments.
Figure~\ref{fig:identification-example} illustrates the mismatch on an
actual question. The cubby scene expresses tiling through repeated
spatial units. The target story uses a repeated sentence template with
varying participants and objects. Gemini 3.8 Flash correctly identifies
these forms and the motifs of the alternative stories in separate
questions, but directly selects the alternative story.
The correct story B and the scene share the motif label tiling, yet the
model's matching answer selects a different motif. The example makes the
distinction between identification and matching directly observable.

\subsection{Model size}

\textbf{Identification improves more consistently than matching as model size increases.}
The Qwen3-VL family provides a comparison across model sizes under the
same task prompts. Scene and story identification improve across the
evaluated variants. Scene-to-story matching follows a different profile,
as shown in Figure~\ref{fig:main-findings}c. The larger variants improve
identification while matching fluctuates and returns to the accuracy
of a smaller variant.

The within-family trend complements the paired experiment. Successful
identification can coexist with matching errors on the same questions,
and stronger identification across model sizes need not yield stronger
matching.

\subsection{Target voice}

\textbf{Scenes are generally easier to match to mathematical statements than to stories.}
The scene-to-story and scene-to-mathematical-statement tasks use the same
source-scene pool and the same number of candidates. Most evaluated
models achieve higher matching accuracy with mathematical statements,
as shown in Table~\ref{tab:main}. This includes the models with the
strongest overall and scene-to-story performance.

The target voices express their motifs differently. Mathematical
statements make selected relations explicit through notation or a named
association. Stories express them through the organization of a telling,
requiring the model to distinguish that organization from its subject
matter. The observed difference therefore identifies story form as a
demanding target for matching.

\subsection{Visual presentation}

\textbf{Identification is less reliable in cluttered photorealistic scenes.}
The presentation experiment varies rendering style, scene content, and
motif visibility across a progression from skeletons to cluttered
photorealistic scenes. Figure~\ref{fig:main-findings}d shows the complete
accuracy profiles, averaged within motif and then across motifs.
Every displayed model performs worse at the cluttered endpoint than
at the skeleton endpoint, although intermediate changes are not
monotonic. The task still asks for the same motif labels, showing
sensitivity to how their structural evidence is presented.

Additional matching procedures are reported in
Section~\ref{app:label-matching}.

%% file: revision_storyline/tables/main_results.tex
\begingroup
\setlength{\tabcolsep}{3.2pt}
\renewcommand{\arraystretch}{1.02}
\begin{tabular*}{\linewidth}{@{\extracolsep{\fill}}lccccccccc@{}}
\toprule
Model & Scene & Skeleton & Story & Math & Sc$\to$St & Sc$\to$Math & St$\to$Math & \makecell{Paired\\scene ID} & Avg. \\
\midrule
\multicolumn{10}{@{}l}{\vendorheading{OpenAI}} \\[-1pt]
\modelicon{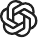}{GPT-5.5} & 77.6\% & 83.7\% & 90.9\% & 88.0\% & \textbf{71.9\%} & 82.4\% & 82.6\% & \textbf{88.1\%} & 83.2\% \\
\modelicon{openai}{GPT-5.2} & 72.9\% & 77.6\% & 76.9\% & 80.0\% & 43.8\% & 63.8\% & 66.1\% & 81.9\% & 70.4\% \\
\addlinespace[1.5pt]
\multicolumn{10}{@{}l}{\vendorheading{Alibaba}} \\[-1pt]
\modelicon{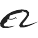}{Qwen3-VL-32B} & 60.5\% & 81.6\% & 63.6\% & 80.0\% & 39.5\% & 54.8\% & 53.7\% & 81.4\% & 64.4\% \\
\modelicon{alibaba}{Qwen2.5-VL-32B} & 50.5\% & 73.5\% & 50.4\% & 76.0\% & 33.3\% & 47.6\% & 57.0\% & 72.9\% & 57.6\% \\
\modelicon{alibaba}{Qwen2.5-VL-72B} & 51.4\% & 61.2\% & 45.5\% & 80.0\% & 38.6\% & 47.1\% & 55.4\% & 76.2\% & 56.9\% \\
\modelicon{alibaba}{Qwen3-VL-8B} & 50.0\% & 69.4\% & 38.8\% & 68.0\% & 35.2\% & 50.5\% & 47.9\% & 73.8\% & 54.2\% \\
\modelicon{alibaba}{Qwen3-VL-4B} & 42.4\% & 75.5\% & 33.1\% & 68.0\% & 39.5\% & 47.6\% & 43.0\% & 71.0\% & 52.5\% \\
\modelicon{alibaba}{Qwen3-VL-2B} & 31.0\% & 61.2\% & 24.0\% & 40.0\% & 28.6\% & 32.4\% & 28.9\% & 62.9\% & 38.6\% \\
\modelicon{alibaba}{Qwen2-VL-7B} & 31.4\% & 44.9\% & 19.8\% & 60.0\% & 28.6\% & 32.9\% & 35.5\% & 55.7\% & 38.6\% \\
\addlinespace[1.5pt]
\multicolumn{10}{@{}l}{\vendorheading{Google}} \\[-1pt]
\modelicon{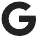}{Gemini 3.8 Flash} & \textbf{86.2\%} & \textbf{95.9\%} & \textbf{91.7\%} & \textbf{92.0\%} & 66.2\% & \textbf{86.7\%} & \textbf{86.8\%} & 87.6\% & \textbf{86.6\%} \\
\modelicon{google}{Gemma-4-31B} & 71.9\% & 83.7\% & 81.0\% & 88.0\% & 44.8\% & 67.1\% & 71.1\% & 82.9\% & 73.8\% \\
\modelicon{google}{Gemma-3-27B} & 46.2\% & 57.1\% & 36.4\% & 64.0\% & 35.7\% & 45.7\% & 43.8\% & 67.6\% & 49.6\% \\
\modelicon{google}{Gemma-3-12B} & 33.3\% & 44.9\% & 36.4\% & 44.0\% & 34.3\% & 42.4\% & 41.3\% & 63.8\% & 42.5\% \\
\addlinespace[1.5pt]
\multicolumn{10}{@{}l}{\vendorheading{Z.AI}} \\[-1pt]
\modelicon{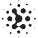}{GLM-4.6V} & 59.0\% & 77.6\% & 50.4\% & 68.0\% & 33.8\% & 50.0\% & 52.9\% & 76.7\% & 58.5\% \\
\modelicon{zhipu}{GLM-4.1V-9B} & 46.2\% & 67.3\% & 47.9\% & 72.0\% & 29.5\% & 51.4\% & 49.6\% & 80.0\% & 55.5\% \\
\modelicon{zhipu}{GLM-4.6V-Flash} & 45.2\% & 73.5\% & 43.0\% & 64.0\% & 33.3\% & 43.3\% & 43.0\% & 72.4\% & 52.2\% \\
\addlinespace[1.5pt]
\multicolumn{10}{@{}l}{\vendorheading{Meta}} \\[-1pt]
\modelicon{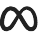}{Llama-4-Scout} & 49.5\% & 65.3\% & 42.1\% & 76.0\% & 31.9\% & 55.7\% & 57.0\% & 73.8\% & 56.4\% \\
\addlinespace[1.5pt]
\multicolumn{10}{@{}l}{\vendorheading{OpenBMB}} \\[-1pt]
\modelicon{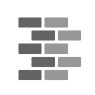}{MiniCPM-V-4.5} & 44.8\% & 59.2\% & 36.4\% & 72.0\% & 34.8\% & 47.6\% & 38.8\% & 67.1\% & 50.1\% \\
\modelicon{openbmb}{MiniCPM-V-4} & 22.4\% & 34.7\% & 14.9\% & 40.0\% & 29.0\% & 37.6\% & 35.5\% & 52.9\% & 33.4\% \\
\addlinespace[1.5pt]
\multicolumn{10}{@{}l}{\vendorheading{OpenGVLab}} \\[-1pt]
\modelicon{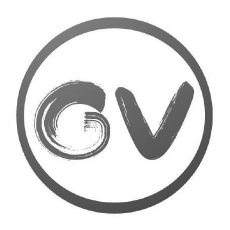}{InternVL3.5-8B} & 33.8\% & 49.0\% & 40.5\% & 68.0\% & 36.2\% & 50.0\% & 57.9\% & 65.7\% & 50.1\% \\
\modelicon{opengvlab}{InternVL3-8B} & 30.5\% & 36.7\% & 33.1\% & 76.0\% & 31.0\% & 42.9\% & 40.5\% & 66.7\% & 44.7\% \\
\addlinespace[1.5pt]
\multicolumn{10}{@{}l}{\vendorheading{Moonshot AI}} \\[-1pt]
\modelicon{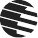}{Kimi-VL-A3B} & 39.5\% & 51.0\% & 24.0\% & 56.0\% & 30.5\% & 37.1\% & 31.4\% & 59.0\% & 41.1\% \\
\addlinespace[1.5pt]
\multicolumn{10}{@{}l}{\vendorheading{Baidu}} \\[-1pt]
\modelicon{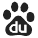}{ERNIE-4.5-VL-28B-A3B} & 31.0\% & 51.0\% & 16.5\% & 36.0\% & 24.8\% & 41.4\% & 37.2\% & 62.9\% & 37.6\% \\
\addlinespace[1.5pt]
\multicolumn{10}{@{}l}{\vendorheading{Mistral AI}} \\[-1pt]
\modelicon{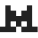}{Pixtral-12B} & 29.0\% & 55.1\% & 36.4\% & 40.0\% & 29.5\% & 39.0\% & 34.7\% & 51.4\% & 39.4\% \\
\addlinespace[1.5pt]
\multicolumn{10}{@{}l}{\vendorheading{NVIDIA}} \\[-1pt]
\modelicon{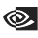}{Nemotron-Nano-VL-8B} & 17.1\% & 26.5\% & 27.3\% & 44.0\% & 32.4\% & 31.4\% & 28.1\% & 58.6\% & 33.2\% \\
\addlinespace[2pt]
Chance & 4\% & 4\% & 4\% & 4\% & 25\% & 25\% & 25\% & 25\% & -- \\
\bottomrule
\end{tabular*}
\endgroup

%% file: revision_storyline/sections/appendix.tex
\section{Construction}\label{app:repro}

\subsection{Generation}
Scenes combine skeleton-conditioned image edits, text-to-image generation
and deterministic repairs. Generation requests use \texttt{gpt-image-2}
for the initial scenes and \texttt{gpt-image-2:stable} for expansion.
Stories use \texttt{gpt-5.5:stable} for writing and
\texttt{gpt-5.2:stable} for screening. These are the recorded request
identifiers; individual generations lack dated backend identifiers.
Skeletons are rendered from coordinate-level construction recipes.

Form reviewers receive the intended story specification. Story
classifiers receive the story, motif menu and disambiguation notes,
with the answer unmarked. Scene checks receive the intended relation.
These construction checks are separate from the label review below.

\subsection{Items and questions}
Table~\ref{tab:motif-counts} gives the item distribution by motif.
Each motif also contributes a mathematical statement for identification.
Item-averaged accuracy weights motifs by these counts, while average
accuracy assigns equal weight to tasks. The scored IDs and display names
are paired in the motif catalog in Table~\ref{tab:catalog}.

\begin{table}[!htbp]
\caption{Unique scenes, skeleton diagrams and stories used in identification.
The totals are 210, 49 and 121. Multiple skeleton variants exist only for
some motifs, and story deduplication leaves one \texttt{strange\_loop}
story. Evaluation uses these unequal class counts.}
\label{tab:motif-counts}
\centering\small
\input{revision_storyline/tables/motif_counts.tex}

\end{table}

Candidate stories contain their complete titles and bodies. Preparation
checks the required form, removes duplicate story bodies and applies
local repairs. Within each candidate set, distractor stories are chosen
to reduce character-length differences from the target while keeping
their full text. Candidate lengths can still differ, so scene-to-story
matching includes both story interpretation and candidate comparison.

\begin{table}[!htbp]
\caption{Identification and matching tasks. Identification uses the full motif
menu except for Family4, whose four candidate identities and order match
the scene-to-story question. Different skeleton renderings can share a
motif. Additional image-withheld and option-rotation settings do not enter
the primary average accuracy.}
\label{tab:tasks}
\centering\small
\input{revision_storyline/tables/tasks.tex}

\end{table}

\section{Evaluation settings}\label{app:evaluation}

\subsection{Models and inference}
The primary GPT runs use \texttt{gpt-5.5-2026-04-23} and
\texttt{gpt-5.2-2025-12-11}. Gemini uses \texttt{gemini-3.8-flash},
and DeepSeek uses \texttt{deepseek-v4-flash-0731}.
The paired study uses \texttt{gpt-5.5}, \texttt{gpt-5.2},
\texttt{gemini-3.8-flash} and \texttt{gemini-2.5-flash}.
Qwen families are described in
\citet{bai2025qwen25vl,wang2024qwen2vl,qwen2025,bai2025qwen3vl}.

Requests use temperature zero. The primary evaluation allows 4,096
output tokens, except OLMo's 1,353-token limit; the paired study allows
8,192 output tokens. Images are resized to a maximum side of 896 pixels
and encoded as JPEG at quality 88. Text-only and image-withheld requests
omit the image field.

\subsection{Scoring}\label{app:answer-scoring}
The primary tasks use compatibility scoring. Answers are extracted from
the requested JSON field, an embedded answer field, or a task-specific
fallback consisting of a motif identifier or the last standalone valid
option letter. The paired study accepts a complete terminal JSON answer,
including a matched code fence. Conflicting answers, unfinished answer
objects and incidental option mentions receive no credit. The same rule
scores each identification answer and its associated matching answer.

Scoring uses the first terminal response. Refusals, exhausted output
budgets and unparseable terminal replies count as incorrect. Requests
without a terminal response remain incomplete and may be retried.
Accuracy uses completed questions as its denominator. Rates, task
averages and differences are computed from integer outcomes before
display rounding. The research package contains the model replies,
scoring code and full rule-comparison outputs.

\subsection{Text-only models}\label{app:extended}
Table~\ref{tab:extmodels} complements the vision-model comparison in the
main text. Its average covers the applicable textual tasks.

\begin{table}[!htbp]
\caption{Primary compatibility accuracy (\%) for all ten text models.
These models are evaluated on three text tasks. Avg. weights those tasks
equally and covers a different task set from the vision average in
Table~\ref{tab:main}.}
\label{tab:extmodels}
\centering\small
\input{revision_storyline/tables/extended_models.tex}

\end{table}

\section{Label review}

\subsection{Model review}\label{app:label-assessment}
We assess all scenes, stories and
skeletons with a model reviewer that receives the item and voice-specific
menu, but neither its intended label nor prior predictions. It may identify
a single best label, several plausible labels, no fitting label or
insufficient evidence. Each plausible interpretation requires localized
visual evidence or a quoted textual relation, with the nearest alternative
and any missing premise. Exact story quotations are checked against the
input text.

Unique agreement counts a single best label equal to the intended one;
set inclusion counts every accepted set containing that label. Multiple
labels, no fitting label and insufficient evidence form separate outcomes.
GPT-6 Astra performs this model review. The primary evaluation
uses the full collection and its construction labels, while supplementary
analyses select items with unique reviewer agreement.

\begin{table}[!htbp]
\caption{Model review. $n$ gives the reviewed items, and all outcome cells are percentages within a voice. The outcome columns form a disjoint partition. Unique match selects the intended label; Other single selects a different label; Multiple accepts several labels; None and Insufficient record the remaining judgments. Rates are computed from exact counts before rounding. Percentages are rounded independently.}
\label{tab:label-evidence}
\centering\small
\input{revision_storyline/tables/label_evidence}

\end{table}

\textbf{Scenes admit a wider range of interpretations than stories.}
The model review assigns unique matching labels less often to scenes than
to stories, as shown in Table~\ref{tab:label-evidence}. The camera scene,
for example, supports self-reference, recursion and nesting under the
revised menu. The outcome categories preserve these differences between
unique agreement and multiple accepted interpretations.

Executable recounting confirms the periodic word-count rule in two
interference stories rejected by the reviewer. Two other reviews quote
genuine text fragments joined by ellipses. These model reviews distinguish the
reviewer's interpretation from the directly checkable story properties.

\subsection{Human evaluation}\label{app:human-evaluation}
Human annotations assess original scene and skeleton images, complete
English stories and mathematical statements through a web interface with
voice-specific definitions. Responses select a unique label, multiple
plausible labels, no suitable label or insufficient evidence.
Table~\ref{tab:human-review} reports completed judgments and agreement
with the intended labels under this assisted protocol.

\begin{table}[!htbp]
\caption{Human evaluation. $n$ is the number of reviewed items; Coverage is their percentage of the available voice pool. Intended, Alternative, Multiple, Insufficient and Inclusion are percentages of the reviewed items. Inclusion is the rate of accepted sets containing the intended label. Model predictions were visible, and some judgments followed intended-label revelation. Rates are computed from per-voice reviewed totals before rounding. Percentages are rounded independently.}
\label{tab:human-review}
\centering\small
\input{revision_storyline/tables/human_review}

\end{table}

Chinese label and definition translations
were added after four submissions; the stimuli and item identifiers
stayed fixed. Per-submission interface versions were not recorded.

Annotations cover the leading portion of each voice's displayed
order. Model predictions were visible throughout, and their labels
identified the menu version used for inference. The intended label was
hidden for 92.0\% of submissions. Within this subset, unique agreement
was 86.1\% and label inclusion was 87.0\%. The remaining 8.0\% followed
intended-label revelation, with 60.0\% unique agreement. These exposure conditions
define the assisted review reported in the paper.

For scenes, the default shortlist contains the Family4 candidates. Other
voices prioritize labels selected by the evaluated models, with
deterministic fillers. The expanded full menu was used for 0.8\% of submissions; the remainder
used four displayed candidates. Shortlists omitted the intended label in
7.2\% of submissions, so those judgments evaluate the displayed alternatives.
Written explanations were optional and none were submitted.

Each reviewed item has one completed submission. The analysis excludes
drafts and interface tests and covers all motif labels, with unequal
coverage across voices and motifs. Agreement rates use this reviewed
subset as their denominator. The original evaluation retains its
construction labels.

\section{Paired experiments}\label{app:paired-experiments}
The paired study compares GPT-5.5, GPT-5.2, Gemini 3.8 Flash and
Gemini 2.5 Flash on shared scene-to-story questions.

\subsection{Identification and matching}
The additional experiment uses the complete scene--story source pool.
Each scene keeps its image, candidate order, intended answer, and all
candidate titles and bodies. The model first identifies the scene among
its four candidate motif labels. Each candidate story is also presented
alone for identification with the same ordered four-label menu.
In matching with the source label, a separate question supplies the
correct source motif label and asks the model to choose among the complete
stories. This condition removes image interpretation while retaining
story reading and selection.

Two menu conditions use the original descriptions and the revised
voice-specific operational descriptions.
The scene and story identification prompts concatenate the common
description with the corresponding voice description.
Matching with the source label supplies the motif ID and common description
under the revised menu, while the original-menu condition supplies the
complete original invariant. Each condition therefore specifies both the
source motif label and the descriptive information available for matching.
The revised menu replaces rhetorical phrasing and unavailable contrast
names with voice-specific descriptions of observable relations. It retains
the same construction labels and marks items with multiple interpretations.

Direct scene--story matching contains no motif menu, so its one response
per source is shared by the two menu analyses. Candidate-story requests
are reused only when their complete text, instruction and ordered menu
are byte-identical. The occurrence map preserves every candidate's
position in the original question.

\subsection{Matching after identification}
We measure matching when the source and target are correctly identified,
and when the source and every candidate are correctly identified.
Each model defines its own subset from its independent identification
answers. Table~\ref{tab:recognition-subsets} reports matching accuracy
alongside coverage of the full paired pool.

\begin{table}[!htbp]
\caption{Direct matching accuracy and subset coverage (\%) under two
identification requirements. The first requires the source and target;
the second requires the source and every candidate story.}
\label{tab:recognition-subsets}\label{tab:decomposition-conditional}
\centering\small
\input{revision_storyline/tables/decomposition_conditional}
\end{table}

\subsection{Label matching}\label{app:label-matching}

\textbf{Label matching improves accuracy over direct matching.}
Direct matching presents the scene and complete candidate stories
together and asks the model to select a story. Label matching first
identifies the scene and each story in separate calls. A deterministic
rule then selects the unique story whose predicted motif label matches
the scene's predicted motif label,
\begin{equation}
B(x,Y)=
\begin{cases}
j,&R_S(x;C)\ne\bot\ \text{and}\
\{k\mid R_T(y_k;C)=R_S(x;C)\}=\{j\},\\
\bot,&\text{otherwise}.
\end{cases}
\label{eq:mediated}
\end{equation}
All labels are predicted by the model. The rule abstains when the source
label is invalid or the matching story is not unique; abstentions count
as errors. Label matching uses additional identification calls, while
the final story selection is performed by the rule. The motif menu
provides names and descriptions for the identification questions.
Matching with the source label is a separate condition that supplies
the correct source motif label and its description.

\begin{table}[!t]
\caption{Accuracy (\%) on the same paired scene-to-story questions.
Source ID and Story ID report identification of the scene and target
story. Direct reports direct matching. Source label reports matching
with the source label. Label matching uses independently predicted motif
labels.
Bold marks the highest accuracy in each column.}
\label{tab:decomposition}
\centering\small
\input{revision_storyline/tables/decomposition}
\end{table}

As shown in Table~\ref{tab:decomposition}, label matching improves accuracy for every model in the paired study.
Label matching converts each item into a motif label
before selecting a candidate. The improvement shows that eliciting the component judgments separately
provides useful answers for the final selection.

Matching with the source label also improves matching accuracy for
each model. It preserves the story candidates and provides the correct
source motif label and its description. This result shows how matching
changes when the source motif is specified. Label matching instead uses
the model's own predictions for both the source and candidate motif
labels.

\subsection{Model review subsets}
Model review subsets require unique reviewer agreement on the source,
on source and target, or on source and every candidate story.
Table~\ref{tab:review-subset-v4} reports their coverage against the full
pool and the corresponding matching accuracies. Construction-branch
summaries use the complete intended-label collection.

\input{revision_storyline/tables/review_subset_v4.tex}

\section{Visual presentation}\label{app:comparison}
The presentation sequence starts with a skeleton (L0), followed by a
sketch (L1), flat-color diagram (L2), minimal vector illustration (L3),
illustration in a simple setting (L4), photo-like scene (L5) and cluttered
photorealistic scene (L6). L0 contains one item per motif; later levels
contain three variants. The levels jointly vary rendering style, scene
content and motif visibility.

Accuracy is averaged within motif and then across motifs at each level.
Table~\ref{tab:auxiliary-results} reports the complete model roster for
the presentation profiles shown in the main figure.
\input{revision_storyline/tables/aux_ladder}

\section{Motif catalog}

The adjacent-swap device evaluates an ordered composition of
transpositions, tracking the resulting permutations. The natural map from
the braid group to the symmetric group has the pure braid group as its
kernel~\citep[Sections 1.1--1.2]{birman2004braids}. The story predicate
checks the permutation sequence described by the crossings.

The substitution device uses the finite prefix obtained by iterating
$A\mapsto AB$ and $B\mapsto A$. Its infinite Fibonacci-word connection is
consistent with the treatment of Sturmian words
\citep[Chapter 2]{lothaire2002sturmian}. The finite check establishes the
specified prefix and permits repeated local blocks.

Descent uses the least-witness contradiction. A positive-integer witness
would imply a strictly smaller witness of the same kind, contradicting
leastness~\citep[Section 1]{conradDescent}. The device represents both
the witness condition and the rule producing a smaller witness.

A contraction has a constant $0\leq\rho<1$ satisfying
$d(Tx,Ty)\leq\rho\,d(x,y)$ for every pair in its metric space
\citep[Theorem 1.1]{conradContraction}. The finite validator checks all
pairs using the specified state set, metric and transition.

\label{app:catalog}

Table~\ref{tab:catalog} lists all 25 motifs with their signature
equation, mathematical home, and the classical theorem attached to each. The
mathematical associations guide construction, and the mathematical statement voice
presented to models is derived from these entries.

\input{revision_storyline/tables/catalog.tex}

\paragraph{Selected constructions.}
The catalog distinguishes a mathematical association from the finite
relation checked in a story. The label \texttt{braiding} uses an ordered
sequence of adjacent swaps. If $s_i$ exchanges adjacent positions, the
state after $t$ exchanges is
\[
p_t=s_{i_t}\circ\cdots\circ s_{i_1}(p_0).
\]
The map $B_n\to S_n$ sends $\sigma_i$ to $s_i$. For example,
$\sigma_i^2$ has identity permutation and braid exponent sum two.
The evaluated quantity is the sequence of seating permutations and their
composition.

The aperiodic-tiling story uses the substitution
$A\mapsto AB$, $B\mapsto A$. Starting from $A$, six applications give
\[
\texttt{ABAABABAABAABABAABABA}.
\]
Local blocks such as \texttt{ABA} repeat. The test checks this finite
prefix and its sentence-type realization. Aperiodicity of an infinite
extension means the absence of a global translational period, not the
absence of recurring local blocks.

The symmetry-breaking story uses one exceptional sentence; role duality
uses an involutive paired-word exchange. The skeletons express their
relations through different visual constructions. Symmetry breaking
changes arrows around a central defect, while figure--ground uses
complementary binary rows. The operational menu describes these
voice-specific items.

The self-reference entry associates its finite acrostic with the diagonal
lemma. The lemma concerns a first-order theory $T$ extending Robinson
arithmetic and a formula $\varphi(x)$ with one free variable. The acrostic
realizes the selected self-referential relation in text.
For recursion, $C_n$ counts full binary trees with $n$ internal nodes
and Dyck words of length $2n$. The encoding $E(\bullet)=\epsilon$ and
$E(\mathrm{Node}(L,R))=(E(L))E(R)$ gives the bijection by first-return
decomposition.

\section{English story example}
\label{app:english-device-example}
The title and body below form the complete scored input for
\texttt{story:symmetry\_en\_s2}.
\begin{quote}
\textbf{Red Mitten}

1. Alice found a red mitten on the laundry hook and took it to Bob by the rain barrel.\\
2. Marta poured sweet tea into three cups while the cat slept and the children hummed a farm tune.\\
3. Alice found a red mitten on the laundry hook and took it to Bob by the rain barrel.
\end{quote}
For this payload, all four recorded checks pass. They test odd nonempty-line count,
consecutive numbering, absence of the specified layer markers, and mirrored
word-token lists. The first and third lines are identical here; the
validator also permits the specified Alice/Bob exchange in other pairs.
The paired lines make the checked relation directly visible in the input.

\section{Model inputs}
Prompt text and released task IDs are reproduced verbatim.
\label{app:complete-inputs}
Image-withheld requests contain the complete options and JSON instruction
in plain text, with both image fields omitted. Family4 keeps the source
image and ordered candidate motifs and requests a \texttt{motif\_id}. The following prompts are reproduced in full from model
requests; line wrapping and mathematical glyphs are typeset for reading.
The shown JPEGs have the exact bytes used by the recorded image
serialization.

\subsection{Mathematical statement matching}
\input{examples/matching_without_image}
This text-only prompt retains the displayed mathematical attributions.

\subsection{Scene to story matching}
\label{app:actual-scene-story}
The following example reproduces a complete scored scene--story question.
It was selected by prompt length without examining model responses.
\begingroup
\let\gebexamplefigure\figure
\let\endgebexamplefigure\endfigure
\renewenvironment{figure}[1][!htbp]{\gebexamplefigure[!htbp]}{\endgebexamplefigure}
\input{examples/scene_story_actual}
\endgroup
\FloatBarrier

\subsection{Paired scene identification}
\begin{center}
\includegraphics[width=0.63\linewidth]{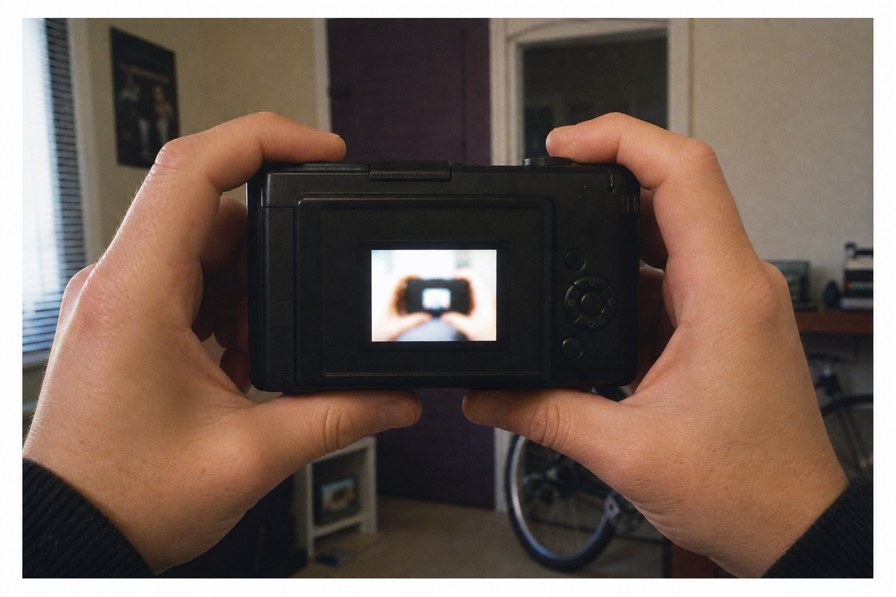}
\end{center}
\input{examples/family4}

\section{Prompts}\label{app:prompts}
The identification and matching templates below give the original question
wording. Substitution fields contain the complete original menu, story title
and body, or candidate statements. Matching candidate blocks use
bracketed A to D labels and full text; motif-identification menus list the
candidate IDs and their descriptions. All templates append the same
single-field JSON instruction appropriate to their answer type.

\begin{longtable}{@{}p{0.24\linewidth}p{0.71\linewidth}@{}}
\caption{Original identification and matching question text before the JSON instruction.
Braced names denote fields replaced with complete input content.}
\label{tab:prompt-templates}\\
\toprule
Task & Question and substituted content \\
\midrule\endfirsthead
\toprule Task & Question and substituted content \\
\midrule\endhead\bottomrule\endfoot
Scene / skeleton & Select the intended structural motif of this image from the menu. Options: \{menu\}. \\
Family4 & Select the intended structural motif of this image from the four options. Options: \{menu4\}. \\
Story & Select the intended structural motif enacted by the form of the story. Story: \{full title and body\}. Options: \{menu\}. \\
Mathematical statement & Select the motif most closely associated with this mathematical statement. Statement: \{text\}. Options: \{menu\}. \\
Scene to story & Select the story whose form most closely matches the intended structure of the image. \{full options\}. \\
Scene to mathematical statement & Select the mathematical statement most closely associated with the intended structure of the image. \{full options\}. \\
Story to mathematical statement & Select the mathematical statement most closely associated with the structure enacted by this story. Story: \{full title and body\}. \{full options\}. \\
Image withheld & No image is available. Select the most plausible answer using only the following options. \{full options\}. \\
\end{longtable}

The appended instruction is \texttt{Output JSON only (no code block):}
followed by \texttt{\{"motif\_id":"..."\}} for identification, including Family4, or
\texttt{\{"choice":"one of A-D"\}} for matching. The strict parser requires
exactly that field with a valid value. The compatibility parser can extract
an embedded field, then a task-appropriate motif identifier or the last
standalone valid option letter. The research package preserves the complete prompts for every evaluation question.

\input{revision_storyline/tables/menu_en.tex}

\paragraph{Menu versions.}
The original descriptions quoted above contain unequal amounts of detail
and some contrast names that are not selectable motif IDs. The new
decomposition experiment uses a separate operational menu. It removes
these contrast names and rhetorical wording, and describes the observable
relation separately for scenes, stories and skeletons. It also identifies
selected constructions such as adjacent-swap composition and the finite
Fibonacci prefix. The paired experiment keeps images, story titles and
bodies, intended keys and candidate order unchanged.

Both menu versions remain available with their corresponding results.
Images and construction labels are shared across the two conditions.
Mathematical statement masking replaces the specified terms while preserving named
attributions and the rest of each statement.

\subsection{Operational menu}
The exact revised descriptions are reproduced in
Table~\ref{tab:menu-v4}. Scene identification uses common plus scene text,
and candidate-story identification uses common plus story text.
The model review uses the description for the displayed voice. The common
text alone accompanies the supplied source motif label in revised-menu
matching with the source label.
\input{revision_storyline/tables/menu_v4.tex}

\section{Skeletons}
\label{app:skeletons}

\begin{figure}[H]
\centering
\includegraphics[width=0.92\linewidth]{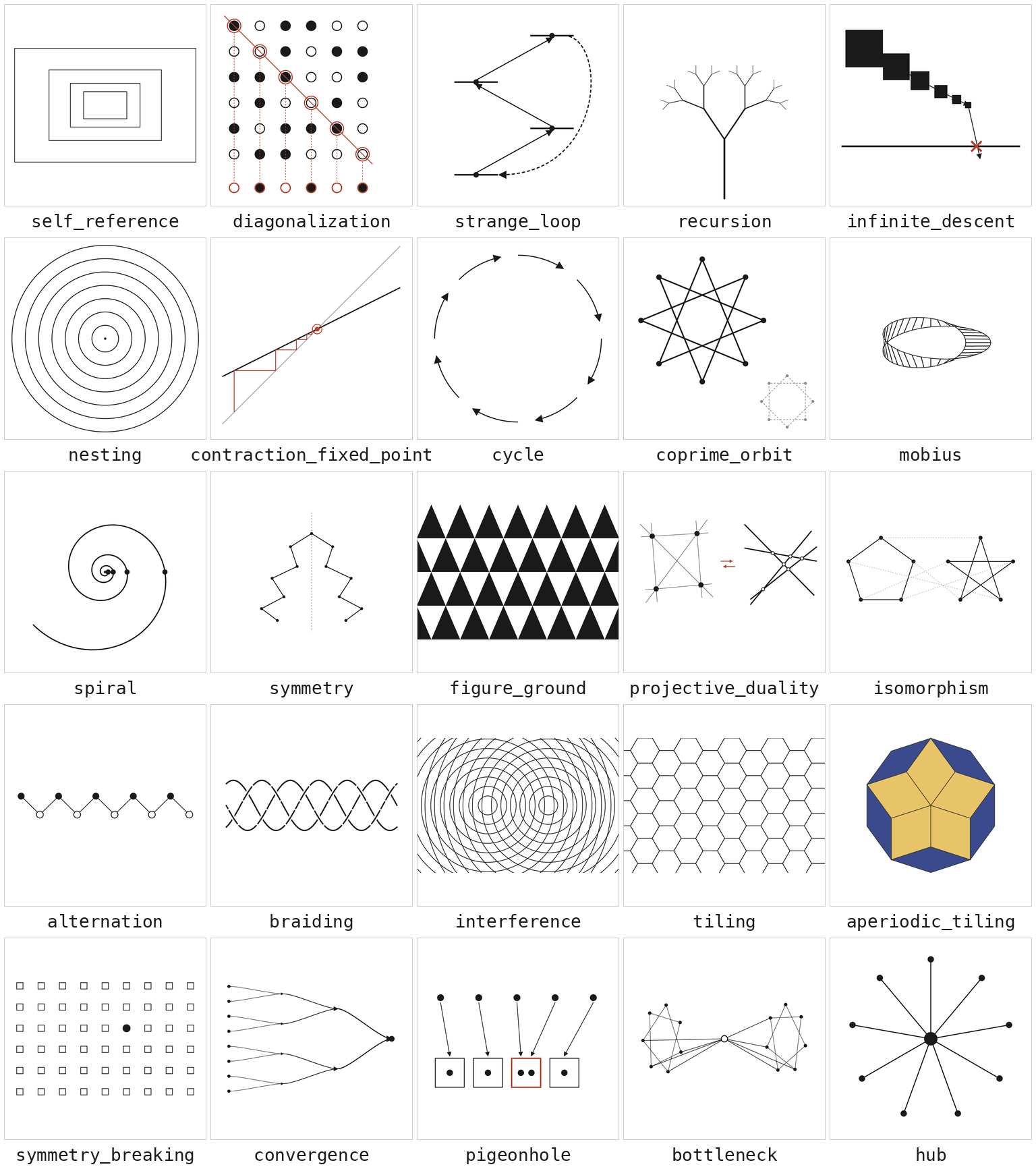}
\caption{All 25 motif skeletons, in the conceptual-arc order of
Figure~\ref{fig:gallery}. Rendering changes count as nuisance variation only when they preserve
the target relation. Item and container counts are essential for
pigeonhole constructions; ring size and stride must retain coprimality
for a single-orbit construction.}
\label{fig:skeleton-gallery}
\end{figure}

Figure~\ref{fig:skeleton-gallery} shows the programmatic skeleton for each of
the 25 motifs as plain 2-D diagrams rendered from coordinate-level recipes.
These images serve as the skeleton-voice identification items and as
references in the skeleton-conditioned generation branch. Their coordinate-level recipes are included in the research package.

\section{Limitations}\label{app:update-limitations}
Our evaluation covers a selection of model families and versions.
The paired experiments compare GPT and Gemini variants, leaving other
architectures and training approaches untested. Broader coverage would
test how consistently the observed identification and matching patterns
generalize across models.

%% file: revision_storyline/tables/motif_counts.tex
\begin{tabular}{lrrr}
\toprule
Motif ID & Scenes & Skeletons & Stories \\
\midrule
self\_reference & 10 & 3 & 5 \\
recursion & 10 & 3 & 5 \\
nesting & 10 & 3 & 5 \\
cycle & 9 & 3 & 5 \\
symmetry & 10 & 1 & 5 \\
alternation & 10 & 3 & 5 \\
spiral & 10 & 3 & 5 \\
convergence & 10 & 1 & 5 \\
bottleneck & 10 & 1 & 5 \\
symmetry\_breaking & 10 & 3 & 5 \\
strange\_loop & 3 & 3 & 1 \\
figure\_ground & 6 & 3 & 5 \\
isomorphism & 10 & 1 & 5 \\
mobius & 10 & 1 & 5 \\
braiding & 8 & 3 & 5 \\
interference & 10 & 1 & 5 \\
tiling & 10 & 3 & 5 \\
hub & 10 & 3 & 5 \\
diagonalization & 5 & 1 & 5 \\
infinite\_descent & 6 & 1 & 5 \\
coprime\_orbit & 10 & 1 & 5 \\
pigeonhole & 5 & 1 & 5 \\
projective\_duality & 8 & 1 & 5 \\
contraction\_fixed\_point & 9 & 1 & 5 \\
aperiodic\_tiling & 1 & 1 & 5 \\
\bottomrule
\end{tabular}

%% file: revision_storyline/tables/tasks.tex
\begin{tabular}{lll r c l}
\toprule
Task & Job & Given $\rightarrow$ pick & Questions & Mode & Released id \\
\midrule
Scene identification    & identify & scene $\rightarrow$ motif    & $210$ & VLM & \texttt{scene2motif} \\
Skeleton identification & identify & skeleton $\rightarrow$ motif & $49$  & VLM & \texttt{skeleton2motif} \\
Story identification    & identify & story $\rightarrow$ motif    & $121$ & LLM & \texttt{story2motif} \\
\makecell[l]{Mathematical statement\\identification}  & identify & mathematical statement $\rightarrow$ motif  & $25$  & LLM & \texttt{theorem2motif} \\
\midrule
Scene--story matching   & match & scene $\rightarrow$ story   & $210$ & VLM & \texttt{crossvoice} \\
\makecell[l]{Scene--mathematical\\statement matching} & match & scene $\rightarrow$ mathematical statement & $210$ & VLM & \texttt{xv\_scene\_thm} \\
\makecell[l]{Story--mathematical\\statement matching} & match & story $\rightarrow$ mathematical statement & $121$ & LLM & \texttt{xv\_story\_thm} \\
\midrule
Paired scene identification  & paired id. & scene $\rightarrow$ motif (of $4$) & $210$ & VLM & \texttt{scene2motif\_family4} \\
\bottomrule
\end{tabular}

%% file: revision_storyline/tables/extended_models.tex
\begingroup
\setlength{\tabcolsep}{3.2pt}
\renewcommand{\arraystretch}{1.02}
\begin{tabular}{@{}lcccc@{}}
\toprule
Model & Story & Math & Story$\to$Math & Avg. \\
\midrule
\multicolumn{5}{@{}l}{\vendorheading{Alibaba}} \\[-1pt]
\modelicon{alibaba}{Qwen2.5-14B} & 45.5 & 72.0 & \textbf{45.5} & 54.3 \\
\modelicon{alibaba}{Qwen3-4B} & 36.4 & 68.0 & 43.0 & 49.1 \\
\modelicon{alibaba}{Qwen2.5-7B} & 26.4 & 64.0 & 41.3 & 43.9 \\
\addlinespace[1.5pt]
\multicolumn{5}{@{}l}{\vendorheading{Google}} \\[-1pt]
\modelicon{google}{Gemma-2-9B} & 29.8 & 60.0 & 36.4 & 42.0 \\
\addlinespace[1.5pt]
\multicolumn{5}{@{}l}{\vendorheading{Z.AI}} \\[-1pt]
\modelicon{zhipu}{GLM-4-9B} & 36.4 & 60.0 & 38.8 & 45.1 \\
\addlinespace[1.5pt]
\multicolumn{5}{@{}l}{\vendorheading{Meta}} \\[-1pt]
\modelicon{meta}{Llama-3.1-8B} & 19.0 & 36.0 & 35.5 & 30.2 \\
\addlinespace[1.5pt]
\multicolumn{5}{@{}l}{\vendorheading{Mistral AI}} \\[-1pt]
\modelicon{mistral}{Mistral-7B-v0.3} & 12.4 & 52.0 & 30.6 & 31.7 \\
\addlinespace[1.5pt]
\multicolumn{5}{@{}l}{\vendorheading{DeepSeek}} \\[-1pt]
\modelicon{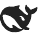}{DeepSeek-V4-Flash-0731} & \textbf{65.3} & \textbf{84.0} & 43.8 & \textbf{64.4} \\
\modelicon{deepseek}{DeepSeek-Coder-V2-Lite} & 16.5 & 40.0 & 27.3 & 27.9 \\
\addlinespace[1.5pt]
\multicolumn{5}{@{}l}{\vendorheading{Allen Institute for AI}} \\[-1pt]
\modelicon{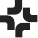}{OLMo-2-7B} & 8.3 & 4.0 & 30.6 & 14.3 \\
\addlinespace[2pt]
Random chance & 4 & 4 & 25 & -- \\
\bottomrule
\end{tabular}
\endgroup

%% file: revision_storyline/tables/label_evidence.tex
\begin{tabular}{@{}lrrrrrr@{}}
\toprule
Voice & $n$ & Unique match & Multiple & Other single & None & Insufficient \\
\midrule
Scene & 210 & 39.0 & 45.2 & 12.9 & 2.9 & 0.0 \\
Story & 121 & 81.8 & 10.7 & 0.0 & 3.3 & 4.1 \\
Skeleton & 49 & 73.5 & 18.4 & 2.0 & 0.0 & 6.1 \\
\bottomrule
\end{tabular}

%% file: revision_storyline/tables/human_review.tex
\begin{tabular}{lrrrrrrr}
\toprule
Voice & $n$ & Coverage (\%) & Intended & Alternative & Multiple & Insufficient & Inclusion \\
\midrule
Scene & 42 & 20.0 & 78.6 & 16.7 & 4.8 & 0.0 & 81.0 \\
Story & 27 & 22.3 & 85.2 & 7.4 & 0.0 & 7.4 & 85.2 \\
Skeleton & 44 & 89.8 & 88.6 & 9.1 & 2.3 & 0.0 & 88.6 \\
\makecell[l]{Mathematical\\statement} & 12 & 48.0 & 83.3 & 16.7 & 0.0 & 0.0 & 83.3 \\
\midrule
Total & 125 & 30.9 & 84.0 & 12.0 & 2.4 & 1.6 & 84.8 \\
\bottomrule
\end{tabular}

%% file: revision_storyline/tables/decomposition_conditional.tex
\begingroup
\setlength{\tabcolsep}{3.5pt}
\renewcommand{\arraystretch}{1.08}
\begin{tabular}{@{}lcccc@{}}
\toprule
Model & \makecell{Source + target\\coverage (\%)} & \makecell{Direct matching\\accuracy (\%)} & \makecell{Source + all stories\\coverage (\%)} & \makecell{Direct matching\\accuracy (\%)} \\
\midrule
\multicolumn{5}{@{}l}{\vendorheading{OpenAI}} \\[-1pt]
\modelicon{openai}{GPT-5.5} & 84.8 & 76.4 & 83.3 & 76.0 \\
\modelicon{openai}{GPT-5.2} & 73.3 & 42.9 & 62.9 & 43.2 \\
\addlinespace[2pt]
\multicolumn{5}{@{}l}{\vendorheading{Google}} \\[-1pt]
\modelicon{google}{Gemini 3.8 Flash} & 89.5 & 71.8 & 89.0 & 71.7 \\
\modelicon{google}{Gemini 2.5 Flash} & 72.9 & 35.9 & 53.8 & 41.6 \\
\bottomrule
\end{tabular}
\endgroup

%% file: revision_storyline/tables/decomposition.tex
\begingroup
\setlength{\tabcolsep}{3.5pt}
\renewcommand{\arraystretch}{1.08}
\begin{tabular*}{\linewidth}{@{\extracolsep{\fill}}lccccc@{}}
\toprule
Model & Source ID & Story ID & \makecell{Source\\label} & Direct & \makecell{Label\\matching} \\
\midrule
\multicolumn{6}{@{}l}{\vendorheading{OpenAI}} \\[-1pt]
\modelicon{openai}{GPT-5.5} & 85.2\% & \textbf{99.5\%} & 87.1\% & \textbf{71.0\%} & 84.3\% \\
\modelicon{openai}{GPT-5.2} & 78.6\% & 94.3\% & 80.5\% & 41.0\% & 70.0\% \\
\addlinespace[2pt]
\multicolumn{6}{@{}l}{\vendorheading{Google}} \\[-1pt]
\modelicon{google}{Gemini 3.8 Flash} & \textbf{91.0\%} & 98.6\% & \textbf{96.2\%} & 67.1\% & \textbf{89.5\%} \\
\modelicon{google}{Gemini 2.5 Flash} & 82.4\% & 89.0\% & 70.5\% & 35.7\% & 66.2\% \\
\bottomrule
\end{tabular*}
\endgroup

%% file: revision_storyline/tables/review_subset_v4.tex
\begingroup
\setlength{\LTcapwidth}{\linewidth}
\setlength{\LTpre}{8pt}
\setlength{\LTpost}{8pt}
\begin{longtable}{@{}llrrr@{}}
\caption{Matching on subsets selected by model review with the revised menu and JSON-only explicit-final scoring. Coverage is the percentage of the full source pool selected by each rule. Direct matching and label matching are accuracy percentages within the selected subset. Selection uses the model reviewer's unique agreement with the intended labels. Bold compares models within each subset and matching method.}
\label{tab:review-subset-v4}
\\
\toprule
Model & Model review subset & Coverage (\%) & Direct (\%) & Label matching (\%)\\
\midrule
\endfirsthead
\multicolumn{5}{l}{\tablename\ \thetable\ (continued)}\\
\toprule
Model & Model review subset & Coverage (\%) & Direct (\%) & Label matching (\%)\\
\midrule
\endhead
\bottomrule\endfoot
\bottomrule\endlastfoot

\modelicon{openai}{GPT-5.5} & Source & 39.0 & \textbf{79.3} & \textbf{97.6} \\
\modelicon{openai}{GPT-5.5} & Source and target & 35.7 & \textbf{82.7} & \textbf{97.3} \\
\modelicon{openai}{GPT-5.5} & Source and all stories & 14.8 & \textbf{83.9} & \textbf{96.8} \\
\addlinespace[2pt]
\modelicon{openai}{GPT-5.2} & Source & 39.0 & 45.1 & 84.1 \\
\modelicon{openai}{GPT-5.2} & Source and target & 35.7 & 45.3 & 86.7 \\
\modelicon{openai}{GPT-5.2} & Source and all stories & 14.8 & 38.7 & 90.3 \\
\addlinespace[2pt]
\modelicon{google}{Gemini 3.8 Flash} & Source & 39.0 & 78.0 & 96.3 \\
\modelicon{google}{Gemini 3.8 Flash} & Source and target & 35.7 & 80.0 & \textbf{97.3} \\
\modelicon{google}{Gemini 3.8 Flash} & Source and all stories & 14.8 & 80.6 & 93.5 \\
\addlinespace[2pt]
\modelicon{google}{Gemini 2.5 Flash} & Source & 39.0 & 40.2 & 78.0 \\
\modelicon{google}{Gemini 2.5 Flash} & Source and target & 35.7 & 40.0 & 80.0 \\
\modelicon{google}{Gemini 2.5 Flash} & Source and all stories & 14.8 & 45.2 & 74.2 \\
\addlinespace[2pt]
\end{longtable}
\endgroup

%% file: revision_storyline/tables/aux_ladder.tex
\begingroup
\setlength{\tabcolsep}{3.2pt}
\renewcommand{\arraystretch}{1.02}
\setlength{\LTcapwidth}{\linewidth}
\setlength{\LTpre}{8pt}
\setlength{\LTpost}{8pt}
\begin{longtable}{@{}lccccccc@{}}
\caption{Ladder accuracy (\%) averaged within motif at each level.
The levels vary rendering style, semantic content and cue visibility.}\label{tab:auxiliary-results}\\
\toprule
Model & L0 & L1 & L2 & L3 & L4 & L5 & L6 \\
\midrule
\endfirsthead
\multicolumn{8}{l}{\tablename\ \thetable\ (continued)}\\
\toprule
Model & L0 & L1 & L2 & L3 & L4 & L5 & L6 \\
\midrule
\endhead
\bottomrule
\endfoot
\bottomrule
\endlastfoot
\multicolumn{8}{@{}l}{\vendorheading{OpenAI}} \\*[-1pt]
\modelicon{openai}{GPT-5.5} & 76.0 & 74.7 & 81.3 & 84.0 & 77.3 & 82.7 & 70.7 \\
\modelicon{openai}{GPT-5.2} & 80.0 & 76.0 & 76.0 & 73.3 & 66.7 & 57.3 & 54.7 \\
\addlinespace[1.5pt]
\multicolumn{8}{@{}l}{\vendorheading{Alibaba}} \\*[-1pt]
\modelicon{alibaba}{Qwen3-VL-32B} & 80.0 & 73.3 & 76.0 & 69.3 & 62.7 & 62.7 & 49.3 \\
\modelicon{alibaba}{Qwen2.5-VL-32B} & 64.0 & 53.3 & 57.3 & 46.7 & 44.0 & 42.7 & 28.0 \\
\modelicon{alibaba}{Qwen2.5-VL-72B} & 56.0 & 52.0 & 53.3 & 52.0 & 45.3 & 45.3 & 34.7 \\
\modelicon{alibaba}{Qwen3-VL-8B} & 60.0 & 58.7 & 56.0 & 56.0 & 42.7 & 42.7 & 33.3 \\
\modelicon{alibaba}{Qwen3-VL-4B} & 56.0 & 50.7 & 49.3 & 50.7 & 42.7 & 40.0 & 25.3 \\
\modelicon{alibaba}{Qwen3-VL-2B} & 52.0 & 52.0 & 52.0 & 48.0 & 33.3 & 33.3 & 24.0 \\
\modelicon{alibaba}{Qwen2-VL-7B} & 44.0 & 40.0 & 40.0 & 32.0 & 30.7 & 33.3 & 21.3 \\
\addlinespace[1.5pt]
\multicolumn{8}{@{}l}{\vendorheading{Google}} \\*[-1pt]
\modelicon{google}{Gemini 3.8 Flash} & \textbf{100.0} & \textbf{92.0} & \textbf{89.3} & \textbf{93.3} & \textbf{92.0} & \textbf{86.7} & \textbf{76.0} \\
\modelicon{google}{Gemma-4-31B} & 84.0 & 78.7 & 74.7 & 81.3 & 74.7 & 72.0 & 65.3 \\
\modelicon{google}{Gemma-3-27B} & 48.0 & 52.0 & 53.3 & 45.3 & 40.0 & 45.3 & 36.0 \\
\modelicon{google}{Gemma-3-12B} & 40.0 & 38.7 & 44.0 & 41.3 & 26.7 & 32.0 & 24.0 \\
\addlinespace[1.5pt]
\multicolumn{8}{@{}l}{\vendorheading{Z.AI}} \\*[-1pt]
\modelicon{zhipu}{GLM-4.6V} & 76.0 & 72.0 & 65.3 & 62.7 & 46.7 & 58.7 & 49.3 \\
\modelicon{zhipu}{GLM-4.1V-9B} & 56.0 & 52.0 & 56.0 & 46.7 & 40.0 & 38.7 & 29.3 \\
\modelicon{zhipu}{GLM-4.6V-Flash} & 68.0 & 61.3 & 60.0 & 56.0 & 42.7 & 41.3 & 32.0 \\
\addlinespace[1.5pt]
\multicolumn{8}{@{}l}{\vendorheading{Meta}} \\*[-1pt]
\modelicon{meta}{Llama-4-Scout} & 56.0 & 61.3 & 65.3 & 56.0 & 38.7 & 46.7 & 28.0 \\
\addlinespace[1.5pt]
\multicolumn{8}{@{}l}{\vendorheading{OpenBMB}} \\*[-1pt]
\modelicon{openbmb}{MiniCPM-V-4.5} & 52.0 & 48.0 & 50.7 & 41.3 & 38.7 & 42.7 & 32.0 \\
\modelicon{openbmb}{MiniCPM-V-4} & 24.0 & 24.0 & 25.3 & 24.0 & 21.3 & 22.7 & 20.0 \\
\addlinespace[1.5pt]
\multicolumn{8}{@{}l}{\vendorheading{OpenGVLab}} \\*[-1pt]
\modelicon{opengvlab}{InternVL3.5-8B} & 36.0 & 37.3 & 32.0 & 28.0 & 29.3 & 24.0 & 21.3 \\
\modelicon{opengvlab}{InternVL3-8B} & 24.0 & 29.3 & 32.0 & 30.7 & 30.7 & 29.3 & 22.7 \\
\addlinespace[1.5pt]
\multicolumn{8}{@{}l}{\vendorheading{Moonshot AI}} \\*[-1pt]
\modelicon{moonshot}{Kimi-VL-A3B} & 40.0 & 45.3 & 46.7 & 44.0 & 44.0 & 42.7 & 30.7 \\
\addlinespace[1.5pt]
\multicolumn{8}{@{}l}{\vendorheading{Baidu}} \\*[-1pt]
\modelicon{baidu}{ERNIE-4.5-VL-28B-A3B} & 36.0 & 38.7 & 37.3 & 29.3 & 24.0 & 26.7 & 22.7 \\
\addlinespace[1.5pt]
\multicolumn{8}{@{}l}{\vendorheading{Mistral AI}} \\*[-1pt]
\modelicon{mistral}{Pixtral-12B} & 48.0 & 36.0 & 42.7 & 38.7 & 22.7 & 24.0 & 14.7 \\
\addlinespace[1.5pt]
\multicolumn{8}{@{}l}{\vendorheading{NVIDIA}} \\*[-1pt]
\modelicon{nvidia}{Nemotron-Nano-VL-8B} & 28.0 & 20.0 & 24.0 & 17.3 & 14.7 & 20.0 & 17.3 \\
\end{longtable}
\endgroup

%% file: revision_storyline/tables/catalog.tex
\begin{center}\scriptsize
\setlength{\tabcolsep}{3pt}
\begin{longtable}{p{2.2cm}p{4.0cm}p{2.9cm}p{3.9cm}}
\caption{The full motif catalog: signature equation, classical
mathematical home, and a gloss of the story form device.}
\label{tab:catalog}\\
\toprule
Motif & Signature & Home & Story device \\
\midrule
\endfirsthead
\toprule
Motif & Signature & Home & Story device \\
\midrule
\endhead
\bottomrule
\endfoot
Self-Reference\newline\texttt{self\_\allowbreak{}reference} & $\vdash\ \psi \leftrightarrow \varphi(\ulcorner \psi \urcorner)$ & Mathematical logic: diagonal lemma; computability: Kleene's second recursion theorem & Acrostic fixed point: the title asserts the acrostic; sentence initials spell the title. \\[2pt]
Recursion\newline\texttt{recursion} & $T ::= \bullet \mid \mathrm{Node}(T,T),\qquad C_n=\tfrac{1}{n+1}\binom{2n}{n}$ & Combinatorics: Catalan numbers and Dyck words; theory of computation: structural induction and the call stack & Nested sub-tales under fixed enter/exit formulas (``now the tale tells'' / ``and back we come''), two children or none, shrinking to a fixed base sentence. \\[2pt]
Nesting\newline\texttt{nesting} & $[a_1,b_1]\supsetneq[a_2,b_2]\supsetneq\cdots\;\Longrightarrow\;\bigcap_{n=1}^{\infty}[a_n,b_n]\neq\varnothing$ & Real analysis: Cantor's nested-interval theorem (compactness); formal languages: proper prefixes of the Dyck language (push-only words) & Open-only frames: openers embed ever deeper, no outer name recurs---push without pop. \\[2pt]
Cycle\newline\texttt{cycle} & $v_{1}\to v_{2}\to\cdots\to v_{n}\to v_{1},\qquad \langle r\mid r^{n}=e\rangle\cong\mathbb{Z}/n\mathbb{Z}$ & Graph theory: directed cycle $C_n$; group theory: cyclic group $\mathbb{Z}/n\mathbb{Z}$; discrete dynamics: periodic orbits of functional graphs & Causal ring: ``thereupon''-linked segments whose last result line equals the first verbatim. \\[2pt]
Mirror Palindrome\newline\texttt{symmetry} & $w=w^{R},\qquad (w^{R})_{i}=w_{n+1-i},\qquad R\circ R=\mathrm{id}$ & Combinatorics on words: palindromic factors and rich words; group theory: dihedral group $D_n$ and involutions & Mirrored numbered word-token lists, with optional Alice/Bob exchange for each pair; punctuation is ignored. \\[2pt]
Alternation\newline\texttt{alternation} & $c(i)=i\bmod 2;\qquad \chi(G)\le 2\iff G\ \text{has no odd cycle}$ & Graph theory: bipartite graphs and 2-coloring (K\"onig's characterization); group theory: $\mathbb{Z}/2\mathbb{Z}$ & Two-voice script: turns labelled by two fixed speaker tags in strict parity alternation. \\[2pt]
Spiral\newline\texttt{spiral} & $r=a\,e^{b\theta}$ & Differential geometry / classical curves: logarithmic (equiangular) spiral; discrete dynamics: geometric growth & Enlarging refrain: one refrain opens four English sections of $6/12/24/48$ words. \\[2pt]
Convergence\newline\texttt{convergence} & $\forall v\in V\quad \exists!\; P:\; v \rightsquigarrow r$ & Graph theory: in-trees (anti-arborescences); rewriting: confluence (Church--Rosser); analysis: Banach fixed-point theorem & Three disjoint threads; all names first co-occur at a new place in the final line. \\[2pt]
Bottleneck\newline\texttt{bottleneck} & $uv\notin E:\;\; \min_{\substack{S\subseteq V\setminus\{u,v\}\\ S\ \text{separates}\ u,v}}\lvert S\rvert \;=\; \max\#\{\substack{\text{internally disjoint}\\u\text{--}v\text{ paths}}\}$ & Graph theory: cut vertices and Menger's theorem & Three threads each cross the single-log bridge exactly once, midway, then fan out and continue. \\[2pt]
Symmetry Breaking\newline\texttt{symmetry\_\allowbreak{}breaking} & $\sum_{p\,:\,X(p)=0} \operatorname{ind}_p(X)=\chi(M)$ & Differential topology: Poincar\'e--Hopf and the hairy-ball theorem; equivariant bifurcation theory: pitchfork bifurcation (spontaneous symmetry breaking) & Selected story device with one exceptional sentence in an otherwise regular sequence. \\[2pt]
Strange Loop\newline\texttt{strange\_\allowbreak{}loop} & $\ell(v)=\ell(u)+1\quad\forall\,(u,v)\in E\setminus\{e^{*}\},\qquad e^{*}=(v_{k},v_{0}),\;\ell(v_{k})=k,\;\ell(v_{0})=0$ & Graph theory: topological sorting and minimum feedback arc sets (DAG criterion; the graph skeleton of G\"odel's diagonal construction) & Three narration levels with a back-edge: the innermost hero is the outer narrator; last line = first. \\[2pt]
Figure and Ground\newline\texttt{figure\_\allowbreak{}ground} & $\exists A\;\bigl(A\in\Sigma_1^0\ \wedge\ \overline{A}\notin\Sigma_1^0\bigr)$ & Recursion theory: recursively enumerable sets and their complements (Post 1944) & The same events retold in the same order with the subject role complemented. \\[2pt]
Isomorphism\newline\texttt{isomorphism} & $\varphi(a\cdot b)=\varphi(a)\circ\varphi(b),\qquad \varphi\colon G\to H\ \text{bijective}$ & Group theory / universal algebra: isomorphism; model theory: isomorphism vs.\ elementary equivalence & Two disjoint vocabularies with a bijective dictionary; sentence $k$ maps to sentence $k$. \\[2pt]
Möbius\newline\texttt{mobius} & $M=[0,1]^2/{\sim},\quad (0,t)\sim(1,1-t)$ & Topology: non-orientable surfaces & Two laps; the second repeats the first with orientation words involuted; only the coda closes it. \\[2pt]
Adjacent-swap composition\newline\texttt{braiding} & $\sigma_i\sigma_{i+1}\sigma_i = \sigma_{i+1}\sigma_i\sigma_{i+1}$ & Low-dimensional topology: Artin braid groups & The participant order is recorded before each adjacent swap; the ordered swaps determine the final permutation. \\[2pt]
Interference\newline\texttt{interference} & $\sin\alpha + \sin\beta = 2\sin\tfrac{\alpha+\beta}{2}\,\cos\tfrac{\alpha-\beta}{2}$ & Fourier analysis: beats (sum-to-product); number theory: least common multiple and the Chinese Remainder Theorem & A period-$3$ opener rule and a period-$4$ ending rule coincide only at sentences $12$ and $24$. \\[2pt]
Tiling\newline\texttt{tiling} & $T+\lambda=T\quad\forall\,\lambda\in\Lambda=\mathbb{Z}v_1\oplus\mathbb{Z}v_2$ & Discrete geometry: wallpaper groups (plane crystallographic groups) & Eight instances of one sentence template, three slots swapped, gapless and overlap-free. \\[2pt]
Hub\newline\texttt{hub} & $G\cong K_{1,n}$ & Graph theory: star graph $K_{1,n}$ (complete bipartite) and degree centrality & Five villagers never speak directly; every quoted pair includes the miller. \\[2pt]
The Diagonal Escape\newline\texttt{diagonal\allowbreak{}ization} & $d(n) = 1 - f(n)(n) \;\Rightarrow\; \forall n:\ d \neq f(n)$ & Set theory / computability: Cantor's diagonal argument & $n$ tellers $\times$ $n$ binary traits; the traveler's trait $k$ flips teller $k$'s trait $k$, so it matches no teller. \\[2pt]
Infinite Descent\newline\texttt{infinite\_\allowbreak{}descent} & $\bigl(\forall n\in\mathbb{N}:\; P(n)\Rightarrow\exists\,m<n,\ P(m)\bigr)\;\Longrightarrow\;\forall n\,\neg P(n)$ & Number theory: Fermat's method of infinite descent / well-ordering of the naturals & A universal positive-integer claimant law, a smaller-witness premise, and a least-witness contradiction. \\[2pt]
Coprime Orbit\newline\texttt{coprime\_\allowbreak{}orbit} & $\langle k \rangle = \mathbb{Z}_n \iff \gcd(k,n)=1$ & Elementary number theory / group theory: generators of the cyclic group $\mathbb{Z}_n$; star polygons $\{n/k\}$ (Poinsot 1810) & A lamp at stride $3$ around $8$: roll-call $0,3,6,1,4,7,2,5$, everyone once; a stride-$2$ comparison visits only half the circle. \\[2pt]
Pigeonhole\newline\texttt{pigeonhole} & $|A| > |B| \;\Rightarrow\; \forall f: A\to B,\ \exists\, x\neq y:\ f(x)=f(y)$ & Combinatorics: Dirichlet's pigeonhole principle (Schubfachprinzip, 1834; used in Diophantine approximation) & $13$ guests, $12$ month-slots; the elder asserts a collision before any roll is read; the count confirms it. \\[2pt]
The Involutive Role-Swap\newline\texttt{projective\_\allowbreak{}duality} & $\sigma:\ (a{:}b{:}c)\ \leftrightarrow\ \{ax+by+cz=0\},\qquad \sigma^{2}=\mathrm{id},\qquad \vdash P\ \Leftrightarrow\ \vdash P^{\sigma}$ & Projective geometry: the principle of duality (Gergonne--Poncelet) & Selected paired-word exchange; applying the same exchange twice restores the original telling. \\[2pt]
The Shrinking Return\newline\texttt{contraction\_\allowbreak{}fixed\_\allowbreak{}point} & $d(f(x),f(y))\le \rho\,d(x,y)\ (\rho<1)\ \text{ on complete }(X,d)\;\Rightarrow\;\exists!\,x^{*}:\ f(x^{*})=x^{*},\quad x_{n+1}=f(x_n)\to x^{*}$ & Analysis / dynamical systems: Banach fixed-point theorem and Picard iteration & A chant deletes its back half each round ($17{\to}9{\to}5{\to}3{\to}2{\to}1$) until one word maps to itself. \\[2pt]
Aperiodic Tiling\newline\texttt{aperiodic\_\allowbreak{}tiling} & $\frac{N_{\text{fat}}(n)}{N_{\text{thin}}(n)}\xrightarrow[n\to\infty]{}\varphi=\frac{1+\sqrt{5}}{2}$ & Discrete geometry / tiling theory and quasicrystal mathematics: Penrose tilings (P1 1974; P3 rhombs 1978); pentagrid algebra (de Bruijn 1981) & Sentence types follow a finite prefix of $A{\to}AB$, $B{\to}A$; local blocks may repeat. \\[2pt]
\end{longtable}
\end{center}

%% file: examples/matching_without_image.tex
\begingroup\small\ttfamily\raggedright
No image is available. Select the most plausible answer using only the following options.\par
\par
[A]\par
Kleene's Second \ensuremath{\square} Theorem (1938): for every total computable function f, there exists an index e such that \ensuremath{\varphi}\_e = \ensuremath{\varphi}\_\{f(e)\}; corollary: in any Turing-complete programming language with an acceptable numbering, there exists a program that \ensuremath{\square} (a \ensuremath{\square}).\par
\par
[B]\par
Cantor's \ensuremath{\square} closed interval theorem: If a sequence of bounded closed intervals satisfies [a\_\{n+1\}, b\_\{n+1\}] \ensuremath{\subseteq} [a\_n, b\_n] for every n, then \ensuremath{\bigcap}\_\{n\ensuremath{\ge}1\}[a\_n, b\_n] is nonempty; if, in addition, b\_n \ensuremath{-} a\_n \ensuremath{\to} 0, then the intersection is exactly the singleton \{\ensuremath{\xi}\}, where \ensuremath{\xi} = sup\_n a\_n = inf\_n b\_n.\par
\par
[C]\par
Hairy Ball Theorem (a corollary of Poincaré\textrm{\textendash}Hopf): On every even-dimensional sphere S\textasciicircum{}\{2n\} (\ensuremath{\chi} = 2 \ensuremath{\ne} 0), every continuous tangent vector field has a zero; moreover, on S\ensuremath{^{2}} one can construct a tangent vector field with exactly one zero (pull back a constant vector field on the plane by stereographic projection; the unique zero is at the north pole and has index 2). Thus a zero is unavoidable, and it can be realized as the only zero.\par
\par
[D]\par
Dyck words of length 2n, full binary trees with exactly n internal nodes, and triangulations of a convex (n+2)-gon are three families of objects all counted by the Catalan number C\_n=\textbackslash{}binom\{2n\}\{n\}/(n+1) (Euler\textrm{\textendash}Segner enumeration; see Stanley, Catalan Numbers), and there are explicit bijections between each pair.\par
\par
Output JSON only (no code block): \{"choice":"one of A-D"\}
\par\endgroup

%% file: examples/scene_story_actual.tex
\begin{figure}[p]
\centering
\includegraphics[width=0.47\linewidth]{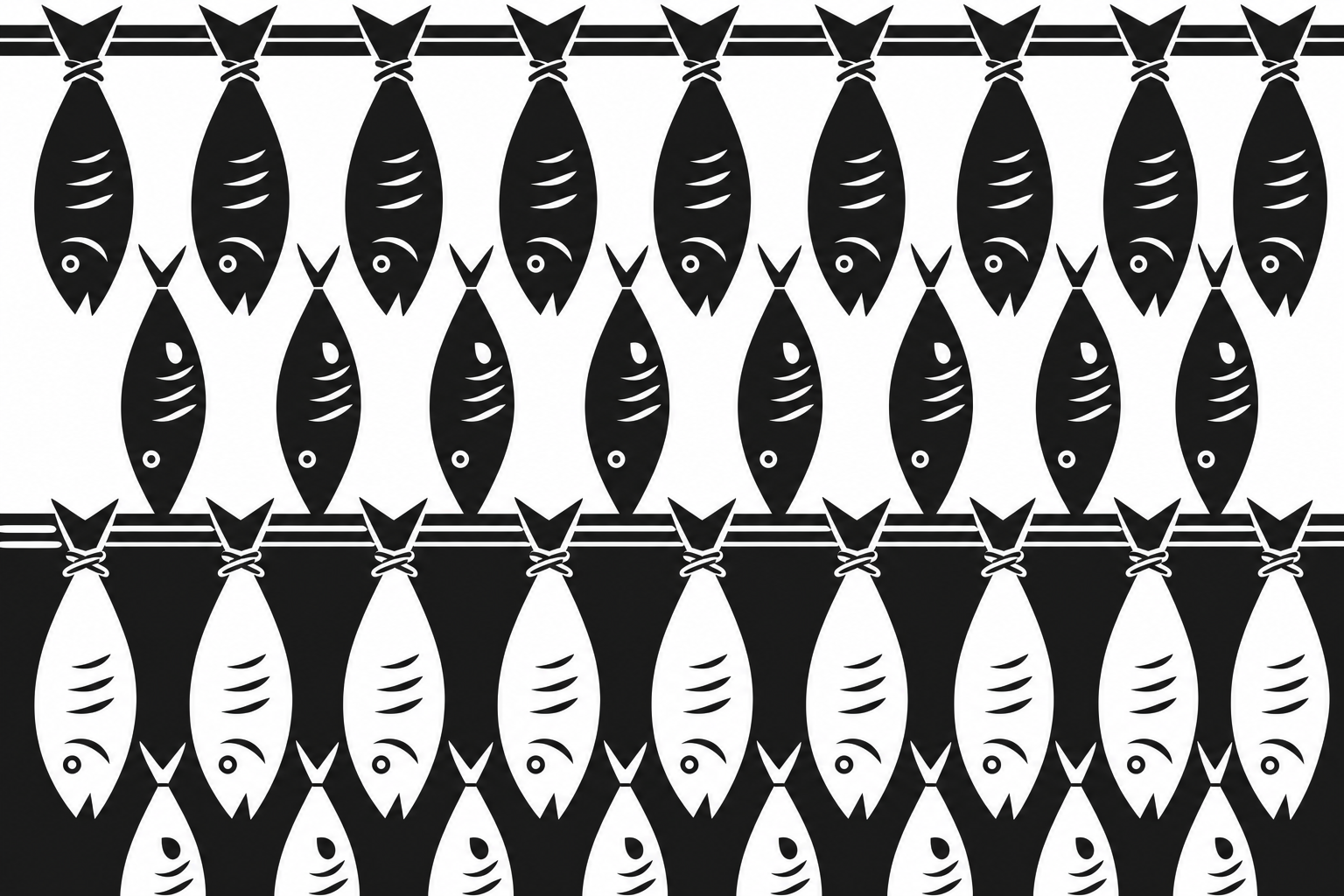}
\caption{An actual scene from a complete scene--story scoring question.
The input and all candidate titles and bodies are reproduced below without
shortening. This example was selected as the shortest complete prompt,
without consulting model performance. The intended answer is D, whose
story exchanges the subject roles across two retellings. The scene's
negative-space interpretation remains a semantic judgment.}
\label{fig:actual-scene-story}
\end{figure}
\begingroup\small
Select the story whose form most closely matches the intended structure of the image.

[A]
Red Mitten

1. Alice found a red mitten on the laundry hook and took it to Bob by the rain barrel.
2. Marta poured sweet tea into three cups while the cat slept and the children hummed a farm tune.
3. Alice found a red mitten on the laundry hook and took it to Bob by the rain barrel.

[B]
The Red Kite

At dawn, Lina carried the red kite left of the shed and smiled. Her brother Tom waved from above the narrow steps and called her down. Inside the tea room, the baker set a warm bun beside the window. The old dog slept below the bench while the cat waited right by the door. At dawn, Lina carried the red kite right of the shed and smiled. Her brother Tom waved from below the narrow steps and called her down. Outside the tea room, the baker set a warm bun beside the window. The old dog slept above the bench while the cat waited left by the door. At dawn, Lina carried the red kite left of the shed and smiled.

[C]
The Hill Gate

Rose $\leftrightarrow$ Lily.
Telling One:
On the hill, Rose carried warm bread to the small gate. Swap once more and the original words return.
Telling Two:
On the hill, Lily carried warm bread to the small gate. Swap once more and the original words return.

[D]
Harbor Light

Alice mended the torn sail for Beth.
Alice shared a bowl of soup with Beth.
Beth mended the torn sail for Alice.
Beth shared a bowl of soup with Alice.

Output JSON only (no code block): \{"choice":"one of A-D"\}
\endgroup

%% file: examples/family4.tex
\begingroup\small\ttfamily\raggedright
Select the intended structural motif of this image from the four options.\par
\par
Options:\par
- self\_reference: The object contains a sentence describing itself, and the description checks out verbatim as true \textrm{\textemdash} it points back to itself in one step, with no climb through levels and no reference to a second system.\par
- nesting: Each time a new level is entered, the previous level is never returned to; depth only ever increases.\par
- symmetry\_breaking: One rule covers the whole field, with exactly one precisely locatable violation; the violation is unique and everywhere else is without exception.\par
- recursion: The whole is generated from smaller copies of itself by the same rule; however many levels you enter, that many you exit, terminating at a base case and then returning level by level along the original path.\par
\par
Output JSON only (no code block): \{"motif\_id":"..."\}
\par\endgroup

%% file: revision_storyline/tables/menu_en.tex
\begin{small}
\begin{longtable}{@{}p{0.22\linewidth}p{0.74\linewidth}@{}}
\caption{The corrected frozen answer menu, reproduced from the administered item file. Wording is part of the model input; global invariant descriptions are not certificates of finite images.}\label{tab:menu}\\
\toprule Motif & Administered description \\
\midrule\endfirsthead
\toprule Motif & Administered description \\
\midrule\endhead\bottomrule\endfoot
\texttt{self\_\allowbreak reference} & The object contains a sentence describing itself, and the description checks out verbatim as true --- it points back to itself in one step, with no climb through levels and no reference to a second system. \\
\texttt{recursion} & The whole is generated from smaller copies of itself by the same rule; however many levels you enter, that many you exit, terminating at a base case and then returning level by level along the original path. \\
\texttt{nesting} & Each time a new level is entered, the previous level is never returned to; depth only ever increases. \\
\texttt{cycle} & Following same-level causal arrows through n\textgreater{}=3 mutually distinct events returns you to the starting event; one lap restores everything verbatim (no escalation, no growth, no flip), all points equal in status, no hierarchy. \\
\texttt{symmetry} & The sequence of events read backwards is item-for-item identical to reading it forwards; folded at the temporal midpoint, it coincides point by point. \\
\texttt{alternation} & Exactly two states strictly take turns: adjacent items always differ, items one apart are always the same, the period is constantly two, without a single exception. \\
\texttt{spiral} & Each time it comes back around to the same bearing it grows one notch by the same fixed ratio, never returning to the original scale. \\
\texttt{convergence} & Multiple threads, mutually unaware, each advance on their own and finally all merge into the same endpoint, and there it ends. \\
\texttt{bottleneck} & Each of the two shores is internally multiply connected, but every path between the shores must pass through one and the same midway point, continuing separately beyond it; deleting that point splits the graph into exactly two still-connected halves, rather than shattering it into isolated pieces. \\
\texttt{symmetry\_\allowbreak breaking} & One rule covers the whole field, with exactly one precisely locatable violation; the violation is unique and everywhere else is without exception. \\
\texttt{strange\_\allowbreak loop} & Climbing step by step up the hierarchy of 'who narrates whom', the final step nevertheless lands back at the starting point; no consistent level can be assigned. \\
\texttt{figure\_\allowbreak ground} & Figure and ground complementarily fill the whole canvas and the ground itself forms a figure: the second half takes the first half's ground as its figure and the first half's figure as its ground, while the narrated events and their order remain unchanged (not reversed order, and not a translation between two vocabularies). \\
\texttt{isomorphism} & Between two things made of entirely different material there exists a one-to-one dictionary preserving all operations and relations, translatable term by term in the same order, without a single exception. \\
\texttt{mobius} & Traveling once around the band returns you to the starting place with orientation flipped (left and right exchanged); only after two laps is everything restored point by point. \\
\texttt{braiding} & Three or more strands cross adjacently several times; the final arrangement of the ends equals exactly the composition of all the crossings in order, and the order cannot be freely exchanged. \\
\texttt{interference} & Two repeating schedules or signals are superposed, producing an interaction pattern. For discrete periods p and q, their joint phase repeats after lcm(p,q); this is a common period, not generally the least period of the sum. The story device checks periods three and four explicitly. \\
\texttt{tiling} & Shift the entire pattern by one cell and it coincides exactly with the original, seamless and without overlap, without a single exception. \\
\texttt{hub} & Between any two endpoints stands only the center point; remove the center and the whole graph shatters into isolated points. \\
\texttt{diagonalization} & A table enumerated row by row; the new row built by flipping each entry along the diagonal is guaranteed to differ from row n at position n, hence differs from every row --- forever outside the table. Unlike self\_reference (an object pointing back to itself in one step): the diagonal element does not talk about itself; the attack targets the table's entire enumeration, not itself. Crucially unlike symmetry\_breaking (exactly one isolated violation in an otherwise regular field): here it is not one anomalous cell but n systematic differences along a moving diagonal --- row n differs at position n, the position shifting right by one each row, at least one specified difference per row, and this construction guarantees 'not equal to any row'; symmetry\_breaking is a single-point flaw that neither escapes nor refutes any enumeration. \\
\texttt{infinite\_\allowbreak descent} & Every case forces a strictly smaller case of the same kind, and the natural numbers admit no bottomless descending chain, so the first case does not exist at all --- the proof's blade swings backwards: not advancing to an end, but cancelling the beginning. Exactly opposite to recursion: recursion exits as many levels as it enters and stands on the existence of a base case; descent stands on the absence of one and condemns every witness to nonexistence. Unlike zeno\_halving: Zeno's steps shrink by a fixed ratio, total length finite, the journey completes; descent's decrease is arbitrary but each step is an integer, the chain is forbidden by well-ordering, the conclusion is 'impossible'. Unlike contraction\_fixed\_point: contraction iterates forever and converges to a genuinely existing fixed point; the descending chain converges to nothing, must hit floor 1 in finitely many steps, so the whole chain together with its starting point is judged nonexistent. Opposite in direction to ordinal\_ascent: ordinal ascent is unbounded upward with every level real; descent is bounded below with the whole chain illusory. Unlike invariant\_barrier: no conserved quantity is invoked; the only killing move is that strict decrease cannot go on forever. \\
\texttt{coprime\_\allowbreak orbit} & On a ring of n points, jump with a fixed stride k, connecting as you go: if the stride is coprime to the ring length, a single stroke visits every point before closing; if not, the ring splits into gcd parts. Unlike cycle (adjacent arrows closing head-to-tail in one lap): the large stride weaves a star shape, traversal is conditional on coprimality, and there is a failure mode --- not coprime means splitting. Unlike interference (two independent periods superposed, an lcm-sized larger period emerging): here there is only one orbit and one number-theoretic condition. Unlike hub (relaying through a center): there is no central point at all; traversal is guaranteed purely by gcd(k,n)=1. \\
\texttt{pigeonhole} & More items than cages forces two into the same cage: the collision is proved certain, yet no one can point to where it happens. Unlike convergence (all threads, knowingly or not, ending at the same KNOWN point): here counting only guarantees that at some UNSPECIFIED place at least two meet; no witness is constructed. Unlike bottleneck (every path forced through the same midway point): there is no forced path and no relay hub, only a single count. \\
\texttt{projective\_\allowbreak duality} & Within a single system, two classes of roles are swapped wholesale and every relational proposition remains true; swapping twice restores the original verbatim (an involution, sigma\textasciicircum{}2=id). Exclusions: unlike isomorphism (a relation-preserving dictionary between two heterogeneous materials) --- duality is a role swap inside one and the same material and is necessarily an involution; unlike figure\_ground (figure and ground complementarily filling one canvas) --- duality produces two pictures each complete in itself, not relying on negative space; unlike transformed\_canon (the same theme reappearing in a lagged voice under a fixed transformation) --- duality is a wholesale sequential restatement that swaps role vocabulary while truth values stay unchanged, not a voice-level pitch/time transformation; unlike chirality (mirror image and original cannot be superposed, the point being 'it does not fit') --- the two dual statements both hold within the same system and the swap restores everything exactly; unlike mobius (the flip requires traveling one full lap along a path) --- duality has no path, it is an immediate wholesale role swap; unlike symmetry (item-by-item reversed palindrome) --- duality does not reverse order, it only swaps roles. \\
\texttt{contraction\_\allowbreak fixed\_\allowbreak point} & The same rule repeatedly transforms its own output toward a state that it leaves unchanged. A strict contraction claim requires a declared complete metric space and a factor q\textless{}1; the story device uses a finite closed chant orbit and an explicitly verified ultrametric. Other images are motif-level illustrations of iterative stabilization. \\
\texttt{aperiodic\_\allowbreak tiling} & Everywhere obeying the same local matching rules, seamlessly covering the whole plane, yet no translation whatsoever makes the whole picture coincide with itself; at the same time any local patch reappears within a bounded gap everywhere --- 'seen before, never repeated'. Exclusion clauses: unlike tiling (shift one cell and the whole coincides; here it is provable that no translational period exists at all); unlike symmetry\_breaking (a periodic background with exactly one locatable violation; here there is no violation anywhere, yet no period exists at all); unlike spiral (phi here is the frequency ratio of the two tile kinds, not level-by-level fixed-ratio growth; the figure has no self-similar scaling center); unlike transformed\_canon (the same theme reappearing verbatim under a fixed transformation; here no block ever recurs exactly at a fixed interval); unlike interference (two periods superposed still yield an lcm period; here the limiting structure's frequency is irrational, so there is never a period). \\
\end{longtable}
\end{small}

%% file: revision_storyline/tables/menu_v4.tex
\begingroup\small
\begin{longtable}{@{}p{0.19\linewidth}p{0.77\linewidth}@{}}
\caption{Complete operational menu used in the additional experiments. Common descriptions define the intended relation; voice descriptions state its selected observable realization. The original motif IDs are unchanged.}\label{tab:menu-v4}\\
\toprule Motif & Operational descriptions\\\midrule\endfirsthead
\toprule Motif & Operational descriptions\\\midrule\endhead
\bottomrule\endfoot
self\_\allowbreak{}reference & \textbf{Common.} A representation refers to the same object or act of representation in which it occurs.\par\smallskip \textbf{Scene.} The represented object or image-making act appears within its own depiction. A written sentence is not required. Similar nested objects alone do not establish this relation.\par\smallskip \textbf{Story.} The title asserts a property of this text, and the sentence initials reproduce that assertion after the specified letter normalization.\par\smallskip \textbf{Skeleton.} A directed edge starts and ends at the same single node. This self-loop is the selected graph realization of self-reference. \\[4pt]
recursion & \textbf{Common.} A construction repeats its own rule within smaller substructures.\par\smallskip \textbf{Scene.} The same structural pattern recurs within smaller parts across levels. A still image can suggest recursive construction but cannot by itself establish a terminating execution and return trace.\par\smallskip \textbf{Story.} Repeated opening and closing sentences form a balanced nesting tree. Each internal story contains smaller stories, and leaves contain an explicit base-case sentence.\par\smallskip \textbf{Skeleton.} A branching tree repeats a smaller parent-child pattern over levels, with filled leaf nodes marking the displayed base cases. The picture alone does not show an execution-return trace. \\[4pt]
nesting & \textbf{Common.} One structure is contained within another through successive levels.\par\smallskip \textbf{Scene.} Objects or regions are successively contained within one another. Containment alone does not establish self-similar construction, execution, or return behavior.\par\smallskip \textbf{Story.} Successive narrators open deeper stories. Earlier narrator and place names do not reappear after entry into the next level, and no closing sentence returns to an outer level.\par\smallskip \textbf{Skeleton.} Successively contained circles have arrows pointing inward and no arrows pointing outward. This shows containment and a directed entry pattern, not an observed temporal execution. \\[4pt]
cycle & \textbf{Common.} A same-level sequence of distinct states reconnects to its starting state.\par\smallskip \textbf{Scene.} A path through at least three distinct positions closes into a ring. Closure must be visible or explicitly indicated rather than inferred from a circular layout alone.\par\smallskip \textbf{Story.} The last sentence of each narrative block repeats the first sentence of the next block. The final block reconnects to the first block, forming one same-level cycle.\par\smallskip \textbf{Skeleton.} Directed edges join distinct nodes into one closed same-level ring, with no other edges. \\[4pt]
symmetry & \textbf{Common.} Parts correspond under reflection or reversal.\par\smallskip \textbf{Scene.} Parts on opposite sides of a reflection axis correspond in arrangement. Approximate scene symmetry and exact geometric symmetry should be distinguished.\par\smallskip \textbf{Story.} Lines on opposite sides of a central line match as word-token sequences, allowing the specified exchange of the two character names. Sentence order is reflected while word order within a line is preserved.\par\smallskip \textbf{Skeleton.} A row of nodes has matching reflected pairs on opposite sides of a central node and a marked reflection axis. \\[4pt]
alternation & \textbf{Common.} Two states alternate in a sequence with period two.\par\smallskip \textbf{Scene.} Two visually distinguishable states alternate along the depicted sequence, with no exception in the visible sequence.\par\smallskip \textbf{Story.} Dialogue lines alternate between two speakers. Every neighboring pair has different speakers, and each speaker recurs after two lines.\par\smallskip \textbf{Skeleton.} Connected nodes alternate between filled and unfilled states along one row, without a visible exception. \\[4pt]
spiral & \textbf{Common.} Repeated turns return to a direction at a progressively different scale.\par\smallskip \textbf{Scene.} A path turns around a center while its radius changes. A fixed growth ratio requires measurable evidence and cannot be inferred from a spiral appearance alone.\par\smallskip \textbf{Story.} A recurring refrain begins separate passages whose word counts repeatedly increase by the same factor while their narrative scope expands.\par\smallskip \textbf{Skeleton.} Successive points wind around a center at increasing radii. Equal angular steps and the declared geometric radius rule define this selected construction. \\[4pt]
convergence & \textbf{Common.} Several initially separate paths merge at a common endpoint.\par\smallskip \textbf{Scene.} Distinct paths visibly merge into the same terminal location rather than merely crossing or passing through a shared intermediate point.\par\smallskip \textbf{Story.} Separately labeled narrative threads keep their entities separate until a shared terminal meeting sentence.\par\smallskip \textbf{Skeleton.} Directed paths merge through intermediate nodes into one final sink with no outgoing edge. \\[4pt]
bottleneck & \textbf{Common.} Connection between two regions requires a shared intermediate passage.\par\smallskip \textbf{Scene.} Every depicted connection between two internally connected regions passes through one intermediate point or passage. Other unseen routes cannot be ruled out from appearance alone.\par\smallskip \textbf{Story.} Distinct journeys each pass through the same intermediate passage and continue beyond it. A closing statement specifies that losing the passage disconnects the two regions.\par\smallskip \textbf{Skeleton.} Two densely connected groups have all between-group paths routed through a shared intermediate node. Removing this node separates the two groups. \\[4pt]
symmetry\_\allowbreak{}breaking & \textbf{Common.} A regular pattern contains a distinguished violation.\par\smallskip \textbf{Scene.} A visible repeated pattern has one identifiable exception. This single-exception construction is the benchmark's selected realization, not a general definition of symmetry breaking.\par\smallskip \textbf{Story.} All but one sentence satisfy the same word-count and keyword rule. The remaining sentence has a different length and omits the keyword.\par\smallskip \textbf{Skeleton.} An arrow field contains a distinguished central defect and changes around that center. The implementation can alter neighboring arrows as well as the center, so it is not a certificate of exactly one differing element. \\[4pt]
strange\_\allowbreak{}loop & \textbf{Common.} A chain that changes representational level returns to its starting entity.\par\smallskip \textbf{Scene.} A depicted hierarchy or change of representational level leads back to the initial entity. An ordinary closed path without level evidence is insufficient.\par\smallskip \textbf{Story.} Successive narrators tell one another's stories across narrative levels. The final block identifies the initial narrator as its protagonist and returns to the opening narrative sentence.\par\smallskip \textbf{Skeleton.} Upward arrows pass through explicitly separated height levels, and a return edge connects the top level to the initial bottom node. Height is the selected level encoding. \\[4pt]
figure\_\allowbreak{}ground & \textbf{Common.} Complementary foreground and background roles can exchange.\par\smallskip \textbf{Scene.} Foreground shapes and their surrounding negative space each form recognizable complementary figures within the same image.\par\smallskip \textbf{Story.} The same events are retold in the same order with two characters exchanging grammatical subject and non-subject roles while the paired main verbs are preserved.\par\smallskip \textbf{Skeleton.} Two aligned rows of binary node states are complementary at every position. This is the selected complement-exchange construction and is not itself a negative-space figure. \\[4pt]
isomorphism & \textbf{Common.} A one-to-one correspondence preserves a specified relation pattern.\par\smallskip \textbf{Scene.} Two distinguishable arrangements admit a one-to-one correspondence preserving the depicted relations. Similar appearance alone is insufficient.\par\smallskip \textbf{Story.} A stated one-to-one word dictionary transforms the first sequence of sentences into the second without changing other words or sentence order.\par\smallskip \textbf{Skeleton.} Two differently arranged graphs have a displayed one-to-one node correspondence preserving the edge relation. \\[4pt]
mobius & \textbf{Common.} A traversal returns to its position after one lap with reversed orientation and restores orientation after two laps.\par\smallskip \textbf{Scene.} A band has a visible half-twist consistent with orientation reversal after one circuit. Occlusion or a crossing can leave this topology ambiguous.\par\smallskip \textbf{Story.} Successive narrative laps apply the same orientation-word exchange. A final repeated sentence demonstrates restoration after two applications.\par\smallskip \textbf{Skeleton.} Two tracks around a ring are joined with a swapped connection so a circuit changes track and a second restores it. The drawn crossing must be distinguished from a graph junction. \\[4pt]
braiding & \textbf{Common.} An ordered sequence of adjacent exchanges determines a final arrangement.\par\smallskip \textbf{Scene.} Several strands or tracked entities undergo successive adjacent exchanges. Endpoint permutation does not uniquely determine braid topology or crossing direction.\par\smallskip \textbf{Story.} A record gives the current order of distinct participants before each adjacent exchange and updates that order after the exchange. The checked result is ordered permutation composition, not a complete braid invariant.\par\smallskip \textbf{Skeleton.} Tracked strands undergo ordered adjacent crossings and finish in the corresponding permuted order. Endpoint order alone does not identify a braid, even if crossing marks are visible. \\[4pt]
interference & \textbf{Common.} Two repeating patterns jointly produce an interaction pattern.\par\smallskip \textbf{Scene.} Two distinguishable repeated structures overlap to produce a combined pattern. A visible overlap alone does not establish a particular least period.\par\smallskip \textbf{Story.} Two textual markers recur according to different specified periods. Their joint occurrences follow the common repetition interval in the displayed sequence.\par\smallskip \textbf{Skeleton.} Two sets of equally spaced lines overlap, with coincident positions at multiples of the joint spacing. \\[4pt]
tiling & \textbf{Common.} A local unit repeats by translation across the represented region.\par\smallskip \textbf{Scene.} A repeated unit covers the depicted region without visible gaps or overlaps. Claims beyond the image require an additional construction rule.\par\smallskip \textbf{Story.} Consecutive sentences share an identical template after the specified variable slots are replaced by their role labels.\par\smallskip \textbf{Skeleton.} A finite region is filled by a repeated hexagonal unit. The target is repeated local placement within the visible region, not unobserved coverage of an infinite plane. \\[4pt]
hub & \textbf{Common.} Peripheral entities connect through one central entity.\par\smallskip \textbf{Scene.} A central node connects to peripheral nodes, and no peripheral-to-peripheral links are depicted. Removing the center leaves the depicted peripheral nodes isolated.\par\smallskip \textbf{Story.} Every recorded exchange involves one central participant, and the prescribed exchanges connect each peripheral participant to this relay.\par\smallskip \textbf{Skeleton.} One central node connects directly to every peripheral node, with no edges between peripheral nodes. \\[4pt]
diagonalization & \textbf{Common.} A new row differs from each listed row at its corresponding diagonal position.\par\smallskip \textbf{Scene.} A displayed binary table and an additional row show the new row disagreeing with row i at column i. A single unusual cell is insufficient.\par\smallskip \textbf{Story.} A table lists binary features for distinct entries. An additional entry's feature at position i flips listed entry i's feature at position i.\par\smallskip \textbf{Skeleton.} An additional binary row flips the highlighted diagonal entry from each corresponding row of a displayed table. \\[4pt]
infinite\_\allowbreak{}descent & \textbf{Common.} A supposed positive-integer witness would require a strictly smaller witness of the same kind, contradicting leastness.\par\smallskip \textbf{Scene.} A depicted descending sequence can illustrate the intended argument, but an image of decreasing size alone does not establish positive-integer witnesses or their impossibility.\par\smallskip \textbf{Story.} A stated law requires every positive whole-number witness to have a smaller witness of the same kind. An explicit least-witness contradiction rejects the supposed initial witness.\par\smallskip \textbf{Skeleton.} Numbered states decrease along arrows toward a positive-integer lower boundary, and a marked impossible continuation crosses that boundary. The image illustrates a least-witness argument rather than providing its full premises. \\[4pt]
coprime\_\allowbreak{}orbit & \textbf{Common.} A fixed modular step visits every position when the step and ring size are coprime.\par\smallskip \textbf{Scene.} A fixed-step path around labeled ring positions visits each position before closing. The ring size and step must be legible to verify coprimality.\par\smallskip \textbf{Story.} A transfer follows a fixed step around labeled ring positions. A coprime-step trial visits every position, whereas a noncoprime-step trial omits a specified subset.\par\smallskip \textbf{Skeleton.} A fixed modular step forms a single orbit through all labeled ring positions, contrasted with a noncoprime step producing separate orbits. Node count and stride are essential. \\[4pt]
pigeonhole & \textbf{Common.} Assigning more items than available containers forces a shared container.\par\smallskip \textbf{Scene.} More distinct items than containers are visibly assigned to containers, forcing at least one shared container. Counts and assignment relations are essential.\par\smallskip \textbf{Story.} More entities than available categories are each assigned to exactly one category. A counting statement guarantees a shared category before the complete assignments are listed.\par\smallskip \textbf{Skeleton.} More item nodes than container nodes are assigned by arrows, and at least one container receives multiple items. Item and container counts are essential. \\[4pt]
projective\_\allowbreak{}duality & \textbf{Common.} A specified exchange of paired roles preserves the displayed relation pattern and is reversed by a second exchange.\par\smallskip \textbf{Scene.} Two depicted relation patterns exchange the specified roles. The selected construction requires an involutive exchange; it does not establish a general projective duality theorem.\par\smallskip \textbf{Story.} A stated paired-word dictionary transforms one telling into the other in the same sentence order. Applying the same exchange again restores the original text.\par\smallskip \textbf{Skeleton.} One displayed point-line configuration is paired with a configuration exchanging point and line roles and incidence counts. The drawing represents a selected dual construction, not a proof that every role exchange is projective duality. \\[4pt]
contraction\_\allowbreak{}fixed\_\allowbreak{}point & \textbf{Common.} Repeated application of one rule reaches or approaches a state unchanged by that rule.\par\smallskip \textbf{Scene.} Successive states visibly approach or reach a stable state. A strict metric contraction cannot be inferred without a declared metric and bound.\par\smallskip \textbf{Story.} A rule repeatedly deletes a specified suffix from a sequence until one word remains and is repeated unchanged. The stated finite metric supports the checked contraction bound.\par\smallskip \textbf{Skeleton.} A cobweb iteration between a function graph and the diagonal approaches their marked intersection. The declared linear function supplies the contraction bound; appearance alone does not. \\[4pt]
aperiodic\_\allowbreak{}tiling & \textbf{Common.} A finite construction follows a substitution rule associated with an aperiodic limit.\par\smallskip \textbf{Scene.} A depicted arrangement is consistent with the declared substitution or matching construction. A finite irregular-looking patch alone cannot establish absence of all translational periods.\par\smallskip \textbf{Story.} Sentence types follow the finite Fibonacci substitution prefix generated by A to AB and B to A. Local blocks may repeat; the finite prefix alone does not prove an infinite sequence aperiodic.\par\smallskip \textbf{Skeleton.} A finite patch combines two rhombus types according to the selected construction. This patch can illustrate an aperiodic tiling system but cannot alone establish absence of a global translational period. \\[4pt]
\end{longtable}
\endgroup